%% file: iclr2024_conference.tex
\documentclass{article} % For LaTeX2e
\usepackage{iclr2024_conference,times}

\input{math_commands.tex}

\usepackage[utf8]{inputenc} % allow utf-8 input
\usepackage[T1]{fontenc}    % use 8-bit T1 fonts
\usepackage{hyperref}       % hyperlinks
\usepackage{url}            % simple URL typesetting
\usepackage{booktabs}       % professional-quality tables
\usepackage{amsfonts}       % blackboard math symbols
\usepackage{nicefrac}       % compact symbols for 1/2, etc.
\usepackage{microtype}      % microtypography
\usepackage{xcolor}         % colors

\usepackage{algorithm}
\usepackage{algorithmic}

\usepackage{amsmath}
\usepackage{graphics}
\usepackage{epsfig}
\usepackage{multirow}
\usepackage{xspace}
\usepackage{subfigure}
\usepackage{fancyhdr}
\usepackage{microtype}
\usepackage{color}
\usepackage{colortbl}
\usepackage{xcolor}
\definecolor{mygray}{gray}{.85}
\definecolor{myyellow}{RGB}{204,102,0}
\definecolor{myred}{RGB}{204,0,102}
\definecolor{mypurple}{RGB}{102,0,204}
\definecolor{maroon}{cmyk}{0,0.87,0.68,0.32}
\definecolor{myblue}{RGB}{227,227,240}

\usepackage{graphics}
\usepackage{amssymb}
\usepackage{amsthm}
\usepackage{mathrsfs}
\usepackage{booktabs}
\usepackage{algorithm}
\usepackage{algorithmic}
\usepackage{arydshln}
\usepackage{siunitx}
\usepackage{appendix}
\usepackage{enumitem}
\usepackage{wrapfig}

\theoremstyle{plain}
\newtheorem{theorem}{Theorem}[section]

\newtheorem{corollary}[theorem]{Corollary}
\theoremstyle{definition}

\theoremstyle{remark}

\newcommand{\model}{NAEPro\xspace}
\newcommand{\layer}{NAEL\xspace}

\usepackage{fancyhdr}
\title{Functional Geometry Guided Protein Sequence and Backbone Structure Co-Design}

\author{%
  Zhenqiao Song$^{1}$, Yunlong Zhao$^{2,3}$, Wenxian Shi$^{4}$, Yang Yang$^{2*}$, Lei Li$^{5}$\thanks{Corresponding author.}\ \\
  $^1$Department of Computer Science, University of California Santa Barbara \\
  $^2$Biomolecular Science and Engineering Program, University of California Santa Barbara\\
  $^3$Department of Chemistry, Massachusetts Institute of Technology  \\
  $^4$Department of EECS, Massachusetts Institute of Technology  \\
  $^5$Language Technology Institute, Carnegie Mellon University\\
  \texttt{\{zhenqiao,yang89\}@ucsb.edu}, \texttt{leili@cs.cmu.edu
}\\
\texttt{\{yunlongz,wxsh\}@mit.edu
} \\}

\iclrfinalcopy % Uncomment for camera-ready version, but NOT for submission.

\begin{document}
\maketitle

\thispagestyle{fancy}            
\fancyhead{}                    
\lfoot{Preprint.}                 
\renewcommand{\headrulewidth}{0pt}
\renewcommand{\footrulewidth}{0pt}
 
\pagestyle{empty}

\begin{abstract}
\input{000abstract}
\end{abstract}

\section{Introduction}
\input{001introduction}

\section{Related Work}
\input{006related_work}

% \section{Background}
% \input{002background}

\section{Proposed Method: \model}
\input{003methods}

\section{Experiments}
\input{004experiments}

\section{Analysis of Proteins Designed by \model}
\input{005analysis}

\section{Conclusion}
\input{007conclusion}


\bibliography{iclr2024_conference}
\bibliographystyle{iclr2024_conference}

%%%%%%%%%%%%%%%%%%%%%%%%%%%%%%%%%%%%%%%%%%%%%%%%%%%%%%%%%%%%
\newpage
\appendix
\input{008appendix.tex}

\end{document}

%% file: math_commands.tex
\usepackage{amsmath,amsfonts,bm}

\def\eqref#1{equation~\ref{#1}}
\def\1{\bm{1}}

\DeclareMathAlphabet{\mathsfit}{\encodingdefault}{\sfdefault}{m}{sl}
\SetMathAlphabet{\mathsfit}{bold}{\encodingdefault}{\sfdefault}{bx}{n}

\DeclareMathOperator*{\argmin}{arg\,min}

%% file: 000abstract.tex
Proteins are macromolecules responsible for essential functions in almost all living organisms. Designing reasonable proteins with desired functions is crucial.
A protein's sequence and structure are strongly correlated and they together determine its function.
In this paper, we propose \model, a model to jointly design Protein sequence and structure based on automatically detected functional and conserved sites. 
\model is powered by an interleaving network of attention and equivariant layers, which can capture global correlation in a whole sequence and local influence from nearest amino acids in three dimensional~(3D) space.
Such an architecture facilitates effective yet economic message passing at two levels.
We evaluate our model and several strong baselines on two protein datasets, $\beta$-lactamase and myoglobin.
Experimental results show that our model achieves the highest binding affinity scores among the top-5, top-10 and top-30 candidates. These findings prove the capability of our model to design functional proteins.
Furthermore, in-depth analysis further confirms our model's ability to generate highly effective proteins capable of binding to their target metallocofactors\footnote{We provide code, data and models on \url{https://github.com/JocelynSong/NAEPro.git}}.

%% file: 001introduction.tex
Proteins are crucial macromolecules in almost all living organisms.
A fundamental problem in protein engineering is designing novel proteins with specific biochemical functions such as catalytic activity \citep{park2006design} and therapeutic efficacy \citep{sasportas2009assessment}.
However, protein function arises from a complex coupling of sequence and structure: atoms that comprise protein structure must satisfy physical and chemical constraints while being ``designable" in the sense that the amino-acid sequence should fold into that structure.
This intricate connection between sequence and structure makes the task of functional protein design exceptionally challenging.

Recently, deep learning methods have witnessed impressive progress on different aspects of protein design~\citep{ding2022protein}. 
A representative class of work are pipeline-based approaches. These approaches typically start by prioritizing the design of the structure~\citep{trippe2022diffusion, yim2023se, lin2023generating, watson2023novo}, followed by the utilization of existing inverse folding models~\citep{ingraham2019generative, jing2020learning} like ProteinMPNN \citep{dauparas2022robust} to determine sequences that can fold into the specified structure.
This kind of methods have been proven that they can generate new protein sequences with desired functions~\citep{sumida2023improving,zhou2023conditional}.
% \textcolor{red}{Despite their great potential for novel structure design, such sequential design policy fails to cross-condition on sequence and structure, which might lead to inconsistent proteins and inefficient design process~\citep{shi2022protein}}. 
% To improve the consistency, 
Another line of methods is to co-design protein sequence and structure through cross-conditioning\citet{shi2022protein} .
Although their approach is applicable to proteins of various topologies, their model relies on prespecified secondary structure to achieve general protein design, which cannot guarantee the designed proteins exhibit the desired functions.  
\citet{wang2022scaffolding} provide a solution to fill in additional sequence and structure given functional motifs.
However, this approach assumes that biologists already possess knowledge of the motifs associated with the target proteins, thereby potentially restricting the model's applicability.
How can we design functional and consistent protein sequence and structure efficiently?

In this paper, we propose \model to jointly design protein sequence and structure guided by automatically detected meaningful protein fragments. 
We are motivated by the established wisdom that a protein's functionality is closely tied to its functionally critical sites, also known as motifs. Hence, by co-designing both elements based on these functional motifs, we can create functional proteins with consistent sequences and structures. 
Specifically, \model is an interleaving network consisting of stacked neighborhood attentive equivariant layers~({\layer}s).
Each \layer is composed of two integral components: a global attention sub-layer and a neighborhood equivariant sub-layer.
The global attention sub-layer aims to capture global correlations across the entire protein sequence to discover favorable amino-acid combinations within one protein family. 
The neighborhood equivariant sub-layer is designed to filter out distant and potentially noisy information, concurrently gathering effective messages from the nearest amino acids in 3D space.
This architectural design facilitates information exchange with varying levels of granularity, promoting a comprehensive interaction among different residues. This enhanced information exchange, in turn, contributes to more consistent and stably folded proteins.
It is worth mentioning that \model updates sequence and structure features of all residues in an one-shot manner, leading to a much more efficient design process.

We carry out experiments on two metalloproteins, including $\beta$-lactamase and myoglobin. The contribution of this paper are listed as follows: 
\begin{itemize}[nosep,leftmargin=2.6em]
    \item We propose \model to jointly design protein sequences and backbone structures. This model is powered by the innovative neighborhood attentive equivariant layers~({\layer}s).
    \item Experiments show that \model achieves highest binding affinity in all cases. Additionally, among the randomly selected 20 cases from top-100 candidates, $100\%$ and $90\%$ of them are highly potential to bind the corresponding metallocofactos, respectively for myoglobin and $\beta$-lactamase. \model can even generate myoglobin sequences with amino acid identity rate as low as $66.0\%$ compared to the closest matches in Uniprot, yet they bear a remarkable resemblance to natural protein structures, with a RMSD of $0.458${\small \AA}.
    \item Our model is at least $17$x faster than all the representative baselines. We will release our datasets, code and models.
\end{itemize}

% \begin{figure}[htbp]
% \begin{minipage}[t]{0.33\linewidth}
% \centering
%  \includegraphics[height=3.0cm]{alpha.pdf}
% \centerline{(a) ESM-2}
% \end{minipage}%
% \begin{minipage}[t]{0.33\linewidth}
% \centering
% \includegraphics[height=3.0cm]{beta.pdf}
% \centerline{(b) Backbone Coordinate}
% \end{minipage}%
% \begin{minipage}[t]{0.33\linewidth}
% \centering
% \includegraphics[height=3.0cm]{gamma.pdf}
% \centerline{(c) Geometry-Guided Decoder}
% \end{minipage}
% 	\caption{Visualization of three proteins of myoglobin, each containing many instances obtained by different crystal methods. Figure~1(a) is the sequence representation from ESM-2~\cite{lin2022language}. Figure~1(b) shows the same proteins obtained by different crystal methods are closer to each other in $3$D space. Figure~1(c) demonstrates protein sequence representations with closer $3$D structure would also be clustered together after being revised by functional geometry.} 
%  \label{figure_latent}
% \end{figure}

%% file: 006related_work.tex
\textbf{Protein Sequence Design}
Protein sequence design has been studied with a wide variety of methods, including traditional directed evolution \citep{arnold1998design,dalby2011strategy,packer2015methods,arnold2018directed} and machine learning methods~\citep{belanger2019biological,angermueller2019model,moss2020boss,terayama2021black}.
Following the success of deep generative models, there are some work focusing on protein sequence design with specific functions, aka. fitness.
They either search satisfactory sequences using deep generative models \citep{brookes2018design,brookes2019conditioning,madani2020progen,kumar2020model,das2021accelerated,hoffman2022optimizing,melnyk2021benchmarking,anishchenko2021novo,ren2022proximal}, or directly generate protein sequences applying deep generative models~\citep{jain2022biological,song2023importance}.
Another class of methods focus on inverse-folding problem~\citep{fleishman2011rosettascripts,ingraham2019generative,xiong2020increasing,mcpartlon2022deep,hsu2022learning},which targets at producing a protein sequence that can fold into a given structure.
Both approaches lack consideration for $3$D structure design, resulting in constrained accuracy and novelty in the design outcomes.
% Neither approach considers designing $3$D structure, leading to limited design accuracy and novelty.

\textbf{3D Protein Design}
\citet{wang2022scaffolding} propose Inpainting to reconstruct both missing protein sequences and structures using provided motifs. However, it's essential to note that this approach necessitates the prior specification of motifs for the target proteins, which requires domain-specific expertise.
Another class of methods focus on novel protein structure design~\citep{trippe2022diffusion,yim2023se,lin2023generating,watson2023novo} and then apply an inverse folding model to identify a sequence based on the given backbone structure, which may not fully utilize the mutual constraints between protein sequence and structure.
\citet{anand2022protein} first propose to co-design protein sequence and structure conditioning on given secondary structures~(SS). Following their work, \citet{shi2022protein} propose to realize general protein design conditioning on SS and binary contact map. However, designing protein relying on its topology cannot guarantee the designed proteins have the desired functions.

%% file: 003methods.tex
% \begin{figure*}
%   \centering
%   \includegraphics[width=8.5cm]{NAEPro_model.pdf}
%   \vspace{-0.6em}
%   \caption{\model architecture, which consists of $L$ stacked neighborhood attentive  equivariant layers~({\layer}s). Each \layer is composed of one global attention sub-layer and one neighborhood equivariant sub-layer.}
%   \label{Fig: model}
% \end{figure*}

A protein consists of a chain of amino acids~(also called residues) connected by peptide bonds, which folds into a proper 3D structure.
Let $\mathcal{A}$ be the set of 20 common amino acids.
We denote the sequence of a $N$-residue protein by $\boldsymbol{s}=\{s_1, s_2, ..., s_N\} \in \mathcal{A}^N$ and their $C_{\alpha}$ coordinates by $\boldsymbol{x}=[\boldsymbol{x}_1, \boldsymbol{x}_2, ..., \boldsymbol{x}_N]^T \in \mathbb{R}^{N\times3}$.
For residue type $s_i\in \mathcal{A}$, where $i \in \{1, 2, ..., N\}$, we denote its one-hot encoding as $\boldsymbol{s}_i=\text{onehot}(s_i)$.
We denote the scaffold, indexed as $\mathcal{F}=\{1, 2, ..., N\}$, within which the meaningful fragment index set is represented by $\mathcal{M}$.

\begin{wrapfigure}[19]{r}{0.6\textwidth}
  \vspace{-12pt}
  \centering
\includegraphics[width=8.3cm]{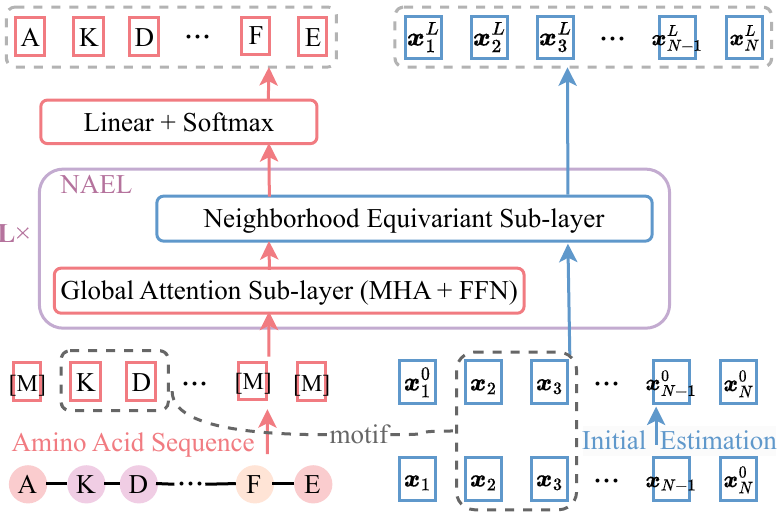}
  \vspace{-12pt}
  \caption{\model architecture, which consists of $L$ stacked neighborhood attentive  equivariant layers~({\layer}s). Each \layer is composed of one global attention sub-layer and one neighborhood equivariant sub-layer.
  } 
  %\vspace{-1.5em}
  \label{Fig: model}
\end{wrapfigure}

The problem studied in this paper can be formulated as follows: given a meaningful fragment index set $\mathcal{M}$, its residue set $\boldsymbol{s}_{\mathcal{M}}$ and its corresponding 3D coordinates $\boldsymbol{x}_{\mathcal{M}}$, generate a protein sequence and $C_{\alpha}$ coordinates of $N$ residues. Essentially, we aim to learn a generative model with probability $P(\boldsymbol{s}, \boldsymbol{x}|\boldsymbol{s}_{\mathcal{M}}, \boldsymbol{x}_{\mathcal{M}})$ where $\boldsymbol{s}$ is a $N$-residue protein sequence and $\boldsymbol{x}$ are their $C_{\alpha}$ coordinates.
The task is challenging because the designed protein sequence $\boldsymbol{s}$ needs to fold into the structure $\boldsymbol{x}$.
Meaningful fragments here represent protein functional and conserved sites, in which functional sites vary from setting to setting. For example, they are active sites for \textit{de novo enzyme design}~\citep{richter2011novo}, while they are binding sites for \textit{de novo binder design}~\citep{gainza2023novo}.

\subsection{Overall Model Architecture}
We propose a model named \model to simultaneously predict the protein sequence $\boldsymbol{s}$ and its 3D backbone structure $\boldsymbol{x}$ given fragment residues $\boldsymbol{s}_{\mathcal{M}}$ and their $C_{\alpha}$ coordinates $\boldsymbol{x}_{\mathcal{M}}$.
\model is a deep neural network consisting of $L$ stacked neighborhood attentive equivariant layers~({\layer}s), which is depicted in Figure~\ref{Fig: model}.
It takes fragment residues and the corresponding coordinates as the input and outputs probabilities of amino acid type for all residues and coordinates of a protein.
Each \layer is composed of a global attention sub-layer using Transformer~\citep{vaswani2017attention} and a neighborhood equivariant sub-layer to incorporate information from nearby residues based on $C_{\alpha}$ coordinates.
Our design rationale centers on computing contextualized representation for each residue while simultaneously enhancing the estimation of each $C_{\alpha}$ coordinate. These two are strongly correlated. Ideally, each residue should receive contextual information from all residues within the protein sequence, as well as from its closest neighbors determined by the 3D structure of the protein.

Suppose $\theta$ are \model parameters. We can formulate the joint sequence and structure probability as:
\begin{equation}
\small
\begin{split}
P(\boldsymbol{s}, \boldsymbol{x}|\boldsymbol{s}_{\mathcal{M}}, \boldsymbol{x}_{\mathcal{M}};\theta)&=P(\boldsymbol{s}|\boldsymbol{s}_{\mathcal{M}}, \boldsymbol{x}_{\mathcal{M}};\theta)\cdot P( \boldsymbol{x}|\boldsymbol{s}_{\mathcal{M}}, \boldsymbol{x}_{\mathcal{M}};\theta) \\
P(\boldsymbol{s}|\boldsymbol{s}_{\mathcal{M}}, \boldsymbol{x}_{\mathcal{M}};\theta)&=\Pi_{i=1\& i\notin \mathcal{M}}^N P(\boldsymbol{s}_i|\boldsymbol{s}_{\mathcal{M}}, \boldsymbol{x}_{\mathcal{M}};\theta), \quad P( \boldsymbol{x}|\boldsymbol{s}_{\mathcal{M}}, \boldsymbol{x}_{\mathcal{M}};\theta) = \Pi_{i=1\&i\notin \mathcal{M}}^N P( \boldsymbol{x}_i|\boldsymbol{s}_{\mathcal{M}}, \boldsymbol{x}_{\mathcal{M}};\theta)\\
P(\boldsymbol{s}_i|\boldsymbol{s}_{\mathcal{M}}, \boldsymbol{x}_{\mathcal{M}};\theta)&=\mathrm{Softmax}(W_{\mathcal{A}} \cdot \boldsymbol{h}_i^{L}),\quad \boldsymbol{x}_i|\boldsymbol{s}_{\mathcal{M}}, \boldsymbol{x}_{\mathcal{M}} \sim \mathcal{N}(\boldsymbol{x}_i^{L}; \lambda I)
\end{split}
\end{equation}
where $W_{\mathcal{A}}$ is the embedding matrix for 20 naturally amino acids, $\boldsymbol{h}_i^{L}$ is the output embedding for $i^{th}$ residue at the last layer of our network and $\boldsymbol{x}_i^{L}$ is the corresponding output coordinate. $\mathcal{N}(\boldsymbol{x}_i^{L}; \lambda I)$ is the Gaussian distribution with mean $\boldsymbol{x}_i^{L}$ and covariance matrix $\lambda I$~($I$ is the identity matrix). $\lambda$ is a hyperparameter.
To find the optimal $\theta$, we maximize the conditional log likelihood~(or equivalent to minimizing the negative log likelihood):
\begin{equation}
\small
\begin{split}
\theta^* &= \argmin_{\theta}\mathcal{L}(\theta)=\argmin_{\theta} -\log P(\boldsymbol{s}, \boldsymbol{x}|\boldsymbol{s}_{\mathcal{M}}, \boldsymbol{x}_{\mathcal{M}};\theta)\\
&=\argmin_{\theta} -\sum\nolimits_{i=1\& i\notin \mathcal{M}}^N \log P(\boldsymbol{s}_i|\boldsymbol{s}_{\mathcal{M}}, \boldsymbol{x}_{\mathcal{M}};\theta)- \sum\nolimits_{i=1\&i\notin \mathcal{M}}^N \log P( \boldsymbol{x}_i|\boldsymbol{s}_{\mathcal{M}}, \boldsymbol{x}_{\mathcal{M}};\theta) 
\end{split}
\end{equation}
For simplicity, we omit the number of protein samples in a dataset. The second log likelihood function can be further simplified as:
\begin{equation}
\small 
\log P( \boldsymbol{x}_i|\boldsymbol{s}_{\mathcal{M}}, \boldsymbol{x}_{\mathcal{M}};\theta) = \log \{\frac{1}{\sqrt{(2\pi)^3}}\exp{(-\frac{1}{2}(\boldsymbol{x}_i -\boldsymbol{x}_i^{L})^T\lambda (\boldsymbol{x}_i -\boldsymbol{x}_i^{L}))}\} = -\frac{\lambda}{2} ||\boldsymbol{x}_i -\boldsymbol{x}_i^{L}||_2^2 + \mathrm{const}
\end{equation}
Therefore, the overall training objective is:
\begin{equation}
\small
\mathcal{L}(\theta) \\
=-\sum\nolimits_{i=1\& i\notin \mathcal{M}}^N \log P(\boldsymbol{s}_i|\boldsymbol{s}_{\mathcal{M}}, \boldsymbol{x}_{\mathcal{M}};\theta) + \frac{\lambda}{2}\sum\nolimits_{i=1\&i\notin \mathcal{M}}^N ||\boldsymbol{x}_i -\boldsymbol{x}_i^{L}||_2^2
\label{equation_all}
\end{equation}

To calculate the prediction probability, we propose the neighborhood attentive equivariant layer~(\layer) to infer non-fragment residue representations and their 3D coordinates based on given meaningful fragment residues and the corresponding positions.
Our key insight is to facilitate thorough information exchange among residues at two levels, global sequence level and local Euclidean distance-based neighborhood.

% \begin{figure*}
%   \centering \includegraphics[width=9.6cm]{message_update.pdf}
%   \vspace{-0.7em}
%   \caption{Neighborhood equivariant sub-layer: neighborhood message update~(green), coordinate update (blue) and residue update (red). Indexed selection is choosing $\boldsymbol{x}_j$ (or $\boldsymbol{h}_j$) where $j^{th}$ residue is in the k-nearest neighbors~(kNN) of $i^{th}$ residue.}
%   \label{Fig: message_update}
% \end{figure*}

\subsection{\layer Global Attention Sub-Layer}
This sub-layer computes global contextual embeddings for all residues. This sub-layer does not consider the closeness of residues in 3D space.
Considering that homologous sequences encode rich semantic features among constituent residues, we allow every residue to attend to all other residues across the whole protein sequence to facilitate information flow through the overall sequence. 
% Through this way, amino-acid combinations frequently occur in the same context would draw higher attention scores. 

We adopt the Transformer layer~\citep{vaswani2017attention} to compute global contextual embeddings. Specifically, each transformer layer is composed of one multi-head self-attention sub-layer~(MHA) and one fully connected feed-forward network~(FFN). A residue connection and a layer normalization are employed after each of the two sub-layers.
Thus, the calculation of global sequence attention can be formulated as follows:
\begin{equation}
\small 
\boldsymbol{h}_i^{l+0.5}=\mathrm{LayerNorm} \left(\mathrm{FFN}(\Tilde{\boldsymbol{h}}_i^{l+0.5})+\Tilde{\boldsymbol{h}}_i^{l+0.5} \right), \quad \Tilde{\boldsymbol{h}}_i^{l+0.5}=\mathrm{LayerNorm}\left(\mathrm{MHA}(\boldsymbol{h}_i^l, \boldsymbol{H}^l)+\boldsymbol{h}_i^l \right)
\end{equation}
where $\boldsymbol{h}_i^{l}$ is the $i^{th}$ residue representation at $l^{\mathrm{th}}$ layer and $\boldsymbol{H}^l=[\boldsymbol{h}_1^l, \boldsymbol{h}_2^l, ..., \boldsymbol{h}_N^l]^T$. 
The input residue embeddings for the first layer are either taken from an embedding lookup table for fragment residues, or initialized with a special [mask] token embedding for non-fragment residues:
\begin{equation} 
\small 
\boldsymbol{h}_i^0 = \left\{
\begin{aligned}
\boldsymbol{s_i}^TW_{\mathcal{A}} &,  & {\text{$i \in \mathcal{M}$, i.e. $i^{th}$ residue is in fragment},} \\
\text{Emb([mask]}) &, & {\text{otherwise}}
\end{aligned}
\right.
\end{equation}
where $W_{\mathcal{A}} \in \mathbb{R}^{20\times d}$ is the embedding matrix for 20 amino acids and $d$ is the dimensionality of residue embeddings. 

\begin{wrapfigure}[18]{r}{0.7\textwidth}
  \vspace{-11pt}
  \centering
\includegraphics[width=9.6cm]{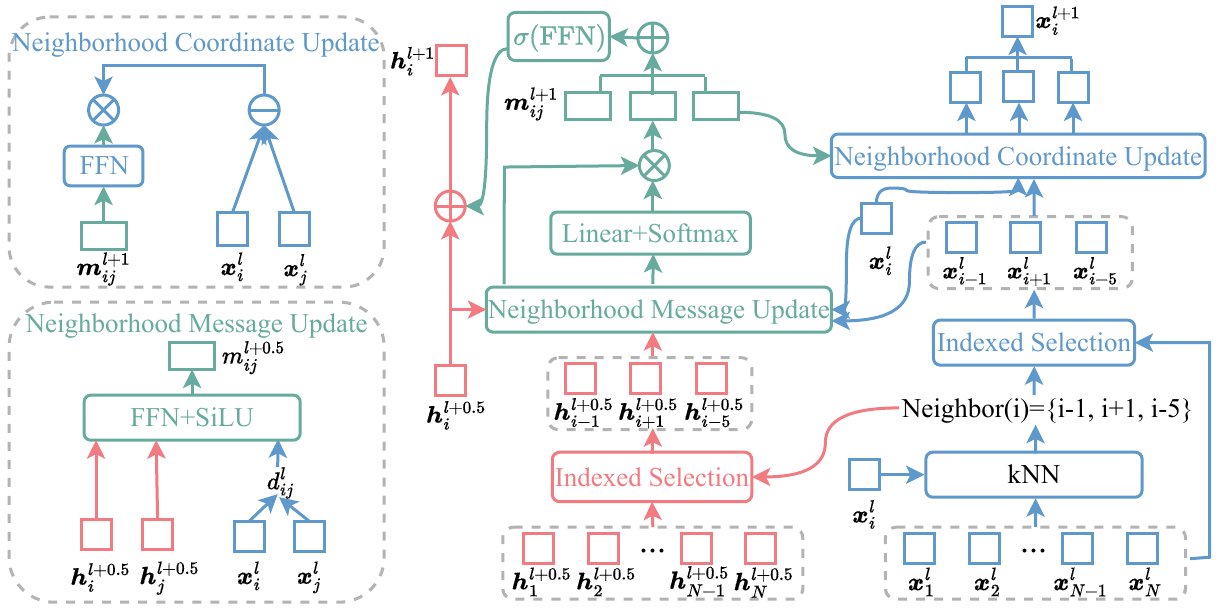}
  \vspace{-9pt}
  \caption{Neighborhood equivariant sub-layer: neighborhood message update~(green), coordinate update (blue) and residue update (red). Indexed selection is choosing $\boldsymbol{x}_j$ (or $\boldsymbol{h}_j$) where $j^{th}$ residue is in the k-nearest neighbors~(kNN) of $i^{th}$ residue.}
  \label{Fig: message_update}
\end{wrapfigure}

\subsection{\layer Neighborhood equivariant sub-layer}
On a protein, a residue's representation and 3D position can be influenced by its nearest neighboring residues. We intend to model such impact with a properly designed subnetwork while keeping the equivariance under 3D translation, rotation and reflection. To this end, we propose the neighborhood equivariant sub-layer. 
This sub-layer includes three components: neighborhood message update, coordinate update and residue update~(Figure~\ref{Fig: message_update}).
It will calculate the updated 3D coordinates of residues based on impact from their neighbors and the updated residue embeddings considering local neighbors.
Specifically, we calculate k-nearest neighbors for each residue based on the Euclidean distances of residue coordinates. 
Updating residue representations and coordinates in 3D space with only nearest neighbors enables more efficient and economic message passing compared to prior approaches which compute messages on the complete pairwise residue graph. 

\textbf{Neighborhood message update.}
We first compute distances among residues using $C_{\alpha}$ coordinates, and select k nearest residues~(Figure~\ref{Fig: message_update}). 
We compute the messages between $i^{th}$ residue and its k-nearest neighbors~(denoted as Neighbor(i)) as follows:
\begin{equation}
\small 
\begin{split}
\boldsymbol{m}_{ij}^{l+0.5} &= \mathrm{SiLU}(\mathrm{FFN}(\mathrm{Concat}(\boldsymbol{h}_i^{l+0.5}, \boldsymbol{h}_j^{l+0.5}, ||\boldsymbol{x}_i^{l}-\boldsymbol{x}^{l}_{j}||_2))) \\
w_{ij}^{l+0.5} &= \frac{\exp (W_a^l\boldsymbol{m}_{ij}^{l+0.5}+b_a^l)}{\sum_{j'\in \mathrm{Neighbor(i)}} \exp (W_a^l\boldsymbol{m}_{ij'}^{l+0.5}+b_a^l)}\\
\boldsymbol{m}_{ij}^{l+1} &= w_{ij}^{l+0.5} * \boldsymbol{m}_{ij}^{l+0.5} \\
\end{split}
\end{equation}
where FFN is a two-layer fully connected feed-forward network with SiLU activation function after its first layer.
Concat is concatenation operator and $||\boldsymbol{x}_i^{l}-\boldsymbol{x}^{l}_{j}||_2$ is the Euclidean distance of $i^{th}$ and $j^{th}$ residue coordinates at $l^{th}$ layer. 
$W_a^l\in \mathbb{R}^{1\times d}$ and $b\in \mathbb{R}$ are the trainable attention parameters of $l^{th}$ layer.

\textbf{Neighborhood coordinate update.} We update the $C_{\alpha}$ coordinate of $i^{th}$ residue as a k-neasrest neighbor vector field in a radial direction (Figure~\ref{Fig: message_update}). 
The $C_{\alpha}$ coordinate $\boldsymbol{x^{l}_i}$ at $l^{th}$ layer of $i^{th}$ residue is updated with the weighted sum of all relative differences $(\boldsymbol{x}_i^{l}-\boldsymbol{x}_j^{l})_{\forall{j\in \mathrm{Neighbor(i)}}}$:
\begin{equation}
\small 
\begin{split}
\boldsymbol{x}^{l+1}_i=\boldsymbol{x}^{l}_i+\sum\nolimits_{j\in \mathrm{Neighbor(i)}}(\boldsymbol{x}^{l}_i-\boldsymbol{x}^{l}_j)\cdot \mathrm{FFN}(\boldsymbol{m}_{ij}^{l+1})
\end{split}
\end{equation}
where FFN is again a two-layer fully connected feed-forward network with SiLU activation function after its first layer.

The input $C_{\alpha}$ coordinates for the whole model are either the given coordinates for fragment residues, or randomly initialized as 3D points on the spherical surface centered at its left residue since the observed Euclidean distances of every neighboring C$_\alpha$ pair are almost the same~(around $r=3.75${\small \AA}):
\begin{equation} 
\small 
\boldsymbol{x}_i^0 = \left\{
\begin{aligned}
\boldsymbol{x}_i &,  & {\text{$i \in \mathcal{M}$, i.e. $i^{th}$ residue is in fragment}}, \\
\boldsymbol{x}_{i-1}^0+r\cdot[\sin{\omega_1}\cos{\omega_2}, \sin{\omega_1}\sin{\omega_2}, \cos{\omega_1}]^T&, & {\text{otherwise}}
\end{aligned}
\right. 
\end{equation}
where $\omega_1$ and $\omega_2$ are sampled from polar coordinate system. $\omega_1$ is the angle to Z-axis sampled from  $\text{Uniform}(0, \pi)$ and $\omega_2$ is the angle to X-axis sampled from $\text{Uniform}(0, 2\pi)$.
% $\omega_1 \sim \text{Uniform}(0, \pi)$ and $\omega_2 \sim \text{Uniform}(0, 2\pi)$.
% i.e. $3.75$\AA\; with $0.013$ variance for $\beta$-lactamase and $3.76$\AA\; with $0.022$ variance for myoglobin:

\textbf{Neighborhood residue update.} We update the $i^{th}$ residue representation by gathering information from its k-nearest neighbors through a gating mechanism~(Figure~\ref{Fig: message_update}):
\begin{equation}
\small 
\begin{split}
\boldsymbol{c_{i}^{l+1}} &= \sum\nolimits_{j\in \mathrm{Neighbor(i)}} \boldsymbol{m}_{ij}^{l+1} \\
\boldsymbol{h_{i}^{l+1}} &= \boldsymbol{h_{i}^{l+0.5}} + \sigma(\mathrm{FFN}(\boldsymbol{c_{i}^{l+1}})) \odot \boldsymbol{c_{i}^{l+1}}
\end{split}
\end{equation}
where FFN is a two-layer fully connected feed-forward network with ReLU activation function after its first layer. $\sigma$ denotes the sigmoid activation function.

We stack $L$ layers of \layer to form the whole model~(Figure~\ref{Fig: model}). 
The output embedding $\boldsymbol{h}^L_i$ and coordinate $\boldsymbol{x}^L_i$ for the $i^{th}$ residue are both from the last layer.
Then the output probability of amino acid type for $i^{th}$ residue is calculated as:
\begin{equation}
\small 
P(s_i=a|\boldsymbol{s}_{\mathcal{M}}, \boldsymbol{x}_{\mathcal{M}})=\frac{\exp(h_{i, a}^o)}{\sum\nolimits_{a'=1}^{20} \exp(h_{i, a'}^o)}, \quad \boldsymbol{h}_i^o=W_{\mathcal{A}}\cdot \boldsymbol{h}_i^L
\end{equation}

\subsection{Meaningful Protein Fragment Mining}
The functionality of a protein is commonly linked to its functional motifs. Consequently, leveraging these functional motifs, we can design proteins with specific functions. In this context, we present a heuristic method for efficient extraction of protein functional and conserved sites. Given that proteins belonging to the same family often share similar functions, they are highly potential to have identical motifs. Therefore, we leverage multiple sequence alignments~(MSAs) to mine meaningful protein fragments, including functional and conserved sites, within the same protein family.
Specifically, we employ the ClustalW2 method~\citep{anderson2011suitemsa} to perform MSAs for each protein family. Subsequently, we designate residue sites as meaningful fragments when they surpass a prespecified $\tau$\% identity threshold, as determined by the MSA results.

\subsection{Analysis on $SE(3)$ Equivariance}
Equivariance plays a critical role in ensuring consistent and predictable performance in the nuisance transformations applied to the input data.
Thus, we analyze the SE(3) equivariance of our model.
\begin{theorem}
\label{theorem_1}
Let R denotes a rotation matrix from SO(3) group and $\boldsymbol{t}\in \mathbb{R}^{3}$ from the translation group. Our \layer is $SE(3)$-equivariant: \small{$\boldsymbol{H^{l+1}}, R\boldsymbol{x^{l+1}}+\boldsymbol{t} = \mathrm{\layer}(\boldsymbol{H}^{l}, R\boldsymbol{x^{l}}+\boldsymbol{t})$}.
\end{theorem}

\begin{corollary}
Parameterizing \model as a composition of L {\layer}s, and taking $l^{th}$ layer output $\boldsymbol{x}^{l+1}$ and $\boldsymbol{H}^{l+1}$ as the ${(l+1)}^{th}$ layer input. \model is SE(3)-equivariant.
\end{corollary}
We provide a formal proof in Appendix~\ref{proof_section_3_5}.

%% file: 004experiments.tex
We conduct extensive experiments to examine the effectiveness of our proposed method. We aim to address the following questions:
\begin{itemize}
[nosep,leftmargin=2.5em]
    \item \textbf{Function~(Q1)}: Do the \model designed proteins exhibit the desired functions?
    \item \textbf{Efficiency~(Q2)}: What is the inference computational complexity?
    % \item \textbf{Function~(Q3)}: Do the \model designed proteins exhibit the desired functions?
\end{itemize}
% We will answer Q1 and Q2 in this section, and leave Q3 for the Analysis section~(Section~\ref{Section:analysis_section}).

\subsection{Datasets}
We evaluate our \model on two metalloproteins: $\beta$-lactamase and myoglobin. $\beta$-lactamase binds zinc ion and myoglobin binds heme. 
These two belong to a large category of proteins, metalloproteins, which are proteins containing a metal ion cofactor.
Metalloproteins play vital roles in a variety of cell functions, such as storage and transport of proteins. 
We thoroughly discuss the significance of these two metalloproteins in Appendix~\ref{reason_for_datasets}. 
To obtain the data, we first extract all proteins in PDB belonging to the two proteins.
Then we extract chain A for $\beta$-lactamase and all chains for myoglobin. We only keep proteins capable of binding metallofactors.
Next we perform length filtering for both proteins, i.e., reserving $\beta$-lactamase longer than $200$ and myoglobin longer than $100$ to make sure the data are reasonable. 
Then we run MSAs to mine fragments.
Finally, we randomly split each dataset into training/validation/test sets with the ratio $8:1:1$.
The data statistics are given in Appendix~\ref{appendix:data_statistics}.

\subsection{Experimental Details}
\textbf{Implementation Details} 
We use $6$ {\layer}s in \model. The global attention sub-layer parameters are initialized with released ESM-2~\citep{lin2022language} parameters~(esm2\_t6\_8M\_UR50D).
The hyperparameter $\lambda/2$ and $k$ are respectively set to $1.0$ and $30$.
The mini-batch size and learning rate are set to $8$ and $5$e-$4$ respectively.
The model is trained for $100$ epochs with $1$ NVIDIA RTX A$6000$ GPU card. 
The sequences are decoded using greedy decoding strategy. 
To make training easier, we employ an annealing training strategy for the first $10$ epochs, which randomly sample ($85\%$ * ($10$ - epoch) / epoch) residues as pseudo fragments and use real fragments after $10$ epochs.
MSA threshold $\tau$ is respectively set to $30$ and $18$ respectively for myoglobin and $\beta$-lactamase as $\beta$-lactamase has three sub-classes and the sequence is highly variable with length ranging from $200$ to over $1,000$.

\textbf{Baseline Models}
We compare the proposed \model against the following representative baselines:
(1) \textbf{Hallucination}~\citep{anishchenko2021novo} 
uses MCMC~\citep{andrieu2003introduction} incorporating a motif constraint into the acceptance score calculation.
(2) \textbf{Inpainting}~\citep{wang2022scaffolding} recovers both sequence and structure based on the given protein segments.
(3) \textbf{SMCDiff}~\citep{trippe2022diffusion}+\textbf{ProteinMPNN}~\citep{dauparas2022robust}: We first apply SMCDiff to design a protein structure based on given motifs and use ProteinMPNN to generate a sequence based on the given structure.
(4) \textbf{PROTSEED}~\citep{shi2022protein} co-designs protein sequence and backbone structure based on secondary structure and binary contact map for general protein design.
(5) \textbf{FrameDiff}~\citep{yim2023se}+\textbf{ProteinMPNN}: Similar to (3) but with a
SE(3) invariant diffusion model to design structure.
(6) \textbf{RFDiffusion}~\citep{watson2023novo}+\textbf{ProteinMPNN}: Similar to (3) but with a different structure design model which is finetuned from the pretrained RoseTTAFold model~\citep{baek2021accurate}. 

To better analyze the influence of different components in our model, we also conduct the following ablation tests:
(7) \textbf{\model-w/o-gate} replaces the gating mechanism in residue updating process with a MLP like in previous graph neural network.
(8) \textbf{\model-w/o-kNN} replaces the k-nearest neighbor graph with complete pairwise residue graph.
(9) \textbf{\model-w/o-MFFN} replaces FFN with the self-attention mechanism in Transformer during message update.
(10) \textbf{\model-w/o-MCFFN} replaces FFN with the self-attention mechanism in message update and uses the attention weight to replace FFN in coordinate update.
(11) \textbf{\model-w-RandomInit} removes ESM2 parameter initialization.
(12) EGNN+ESM2 concatenates EGNN and ESM2, and then is fientuned on the corresponding dataset.

\textbf{Evaluation Metrics}
We calculate the binding affinity scores applying Gnina~\citep{mcnutt2021gnina} to evaluate the function of the designed proteins. Gnina is a docking tool which can assess how well the designed protein can bind to the corresponding metallocofactors. The lower the docking score, the better the binding affinity. We provide top-5, top-10, top-30 mean binding affinity score as well as median binding affinity score.

% (2) \textbf{RMSD} evaluates how close our designed structure is to the target structure.
% (2) \textbf{pLDDT}~\citep{jumper2021highly} provides an overall confidence score that a designed protein sequence can fold into a structure which is similar to natural proteins. We apply ESMFold~\citep{lin2023evolutionary} to calculate pLDDT due to its much higher efficiency than AlphaFold2~\citep{jumper2021highly}.}
% (4) \textbf{TM-score}~\citep{zhang2005tm} evaluates how similar structure predicted by the designed sequence is to the target structure. We use ESMFold to predict the structure of the designed sequence.
% (5) \textbf{Consistency}: We calculate the RMSD between the designed structure and predicted one from the designed protein sequence to see how well the designed sequence can fold into the designed structure.

\begin{table*}[!t]
\small
\centering
\begin{tabular}{llccccc}
\midrule
& Models & Top-5 ($\downarrow$) & Top-10 ($\downarrow$) & Top-30 ($\downarrow$) & Median ($\downarrow$)   \\
\midrule
\multirow{6}{*}{\rotatebox{90}{$\beta$-lactamase}}& Hallucination & -6.98$\pm$0.01&-6.87$\pm$0.02&-6.69$\pm$0.05&-6.29 \\ 
&Inpainting & -9.89$\pm$0.03&-9.54$\pm$0.16&-9.13$\pm$0.43&-7.24 \\
&SMCDiff+ProteinMPNN & -9.10$\pm$0.01&-9.05 $\pm$0.02&-8.98$\pm$0.01&-6.97  \\
 & PROTSEED & -9.88$\pm$0.21&-9.51$\pm$0.41&-9.01$\pm$0.62&-7.31 \\
& FrameDiff+ProteinMPNN & -9.54$\pm$0.03&-9.56$\pm$0.23&-8.89$\pm$0.35&-7.03 \\
& RFDiffusion+ProteinMPNN & -9.87$\pm$0.05&-9.56$\pm$0.23&-9.12$\pm$0.53&-7.51   \\
&\cellcolor{myblue}\model & \cellcolor{myblue}\textbf{\textbf{-10.06$\pm$0.05}} & \cellcolor{myblue}\textbf{-9.79$\pm$0.10} & \cellcolor{myblue}\textbf{-9.39$\pm$ 0.12}  &\cellcolor{myblue}\textbf{-7.66} \\
\midrule
\multirow{6}{*}{\rotatebox{90}{myoglobin}} & Hallucination & -8.18$\pm$ 0.01&-8.07$\pm$0.03&-7.97$\pm$0.23&-7.25  \\ 
&Inpainting & -13.47$\pm$0.02&-13.12$\pm$0.12&-12.31$\pm$0.54&-9.56\\
&SMCDiff+ProteinMPNN & -11.37$\pm$0.03&-11.12$\pm$0.31&-10.87$\pm$0.42&-8.76 \\
& PROTSEED & -13.21$\pm$0.13&-12.89$\pm$0.42&-11.98$\pm$0.52&-10.23  \\
& FrameDiff+ProteinMPNN & -13.13$\pm$0.05& -12.92$\pm$0.16&-12.21$\pm$0.23&-10.08 \\
&RFDiffusion+ProteinMPNN & -13.68$\pm$0.02&-13.03$\pm$0.21&-12.56$\pm$0.43&-10.15  \\
& \cellcolor{myblue}\model & \cellcolor{myblue}$\textbf{-14.12$\pm$0.01}$ & \cellcolor{myblue}$\textbf{-13.85$\pm$0.10}$  & \cellcolor{myblue}$\textbf{-13.06$\pm$0.38}$ & \cellcolor{myblue}$\textbf{-10.74}$ \\
\bottomrule
\end{tabular}
\vspace{-0.6em}
\caption{Model performance on two metalloprotein datasets. The unit system for top-5, top-10, top-30 mean and median binding affinity score is kcal/mol.}
\label{Tab: main_results}
\end{table*}

\subsection{Main Results}
% Table~\ref{Tab: main_results} reports the performance of all models.
\textbf{Q1: \model can generate proteins that have highest binding affinity.}
In particular, Table~\ref{Tab: main_results} shows that our model achieves the lowest docking scores in all cases, which means our \model can better bind the corresponding metallocofactors. Among all the baselines, RFDiffusion+ProteinMPNN is the current SOTA model and has been validated that it can generate proteins with desired functions.
Table~\ref{Tab: main_results} shows that our model achieves even higher binding affinity than this model. 
It demonstrates that our designed proteins are highly potential to exhibit the desired functions.

% \textbf{Q1: \model can generate protein sequences and structures that closely resemble the natural ones, exhibiting its outstanding capability of designing plausible proteins.}
% In particular, Table~\ref{Tab: main_results} shows that our model achieves best performance in most cases, especially AAR, RMSD and TM-score, on which our \model performs best among all the competitors on both datasets.
% It demonstrates that the sequences and structures designed by our model are highly similar to their natural counterparts, giving evidence that both generated backbone structures and sequences are reliable. 
% Significantly, our model attains a TM-score of $0.5692$ on myoglobin, surpassing the threshold of $0.5$. This demonstrates that, in general, the designed myoglobin sequences are highly potential to correctly fold into their target structures.
% Our interpretation is that our \model leverages both global correlations from the whole sequence and fine-grained messages from the nearest neighbors in 3D space, providing a more knowledgeable design environment.  
% While RFDffusion + ProteinMPNN achieves the highest consistency scores (lowest RMSD) on both datasets, it is worth noting that our model can also design consistent proteins, as those with an RMSD below $5${\small \AA} are considered reasonable.

\begin{figure}
\begin{minipage}[t]{0.33\linewidth}
\centering
\includegraphics[width=4.3cm]{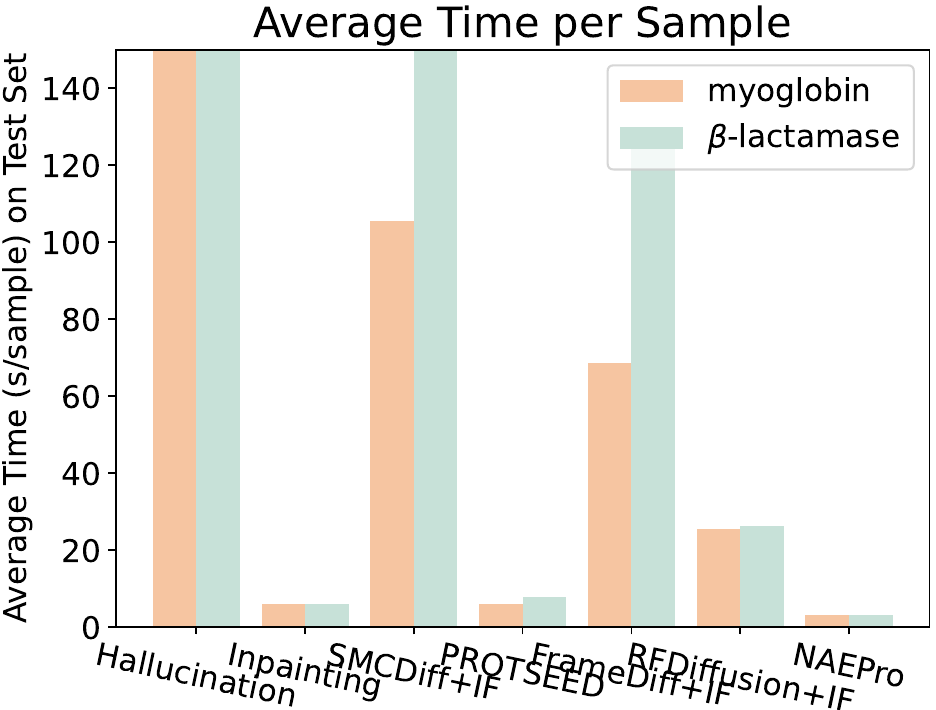}
\centerline{(a) Inference speed}
\end{minipage}%
\begin{minipage}[t]{0.33\linewidth}
\centering
\includegraphics[width=4.3cm]{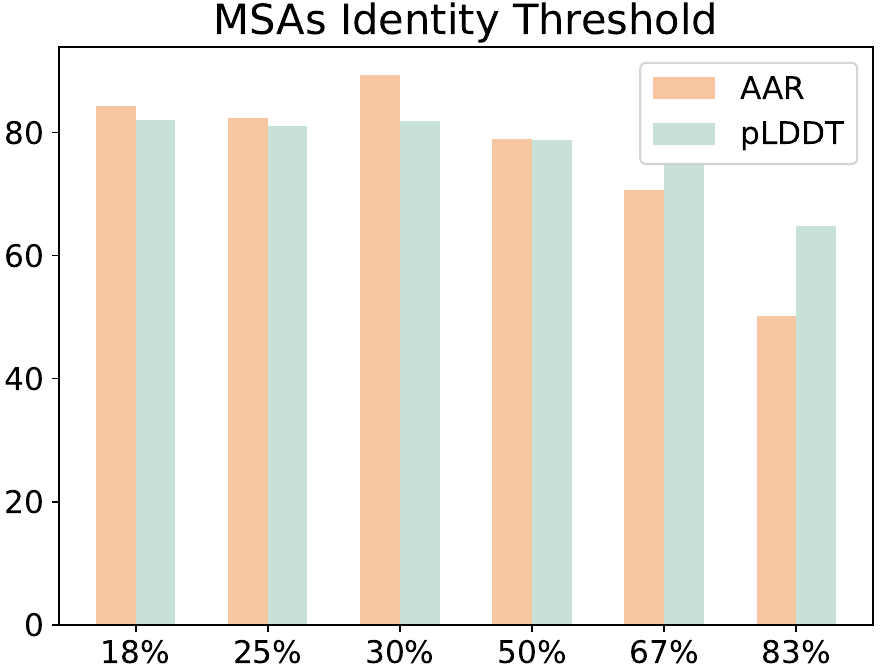}
\centerline{(b) Effects of MSA threshold}
\end{minipage}%
\begin{minipage}[t]{0.33\linewidth}
\centering
\includegraphics[width=4.2cm]{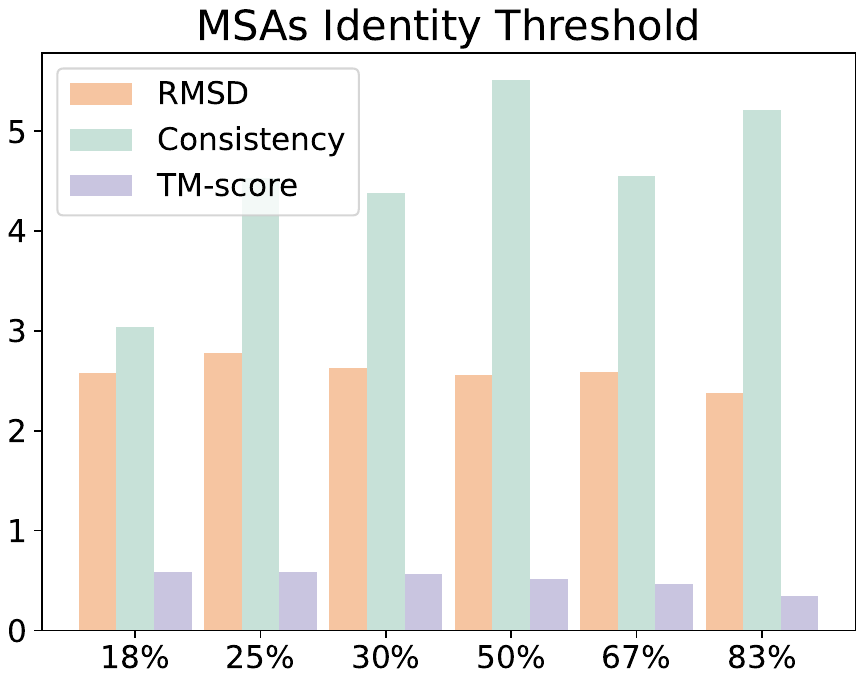}
\centerline{(c) Effects of MSA threshold}
\end{minipage}
	\caption{Visualization of (a) inference speed of all models evaluated by average design time on test set. (b) and (c) model performance on myoglobin under different MSA identity thresholds.} 
 \label{figure_analysis}
\end{figure}

\textbf{Q2: \model inference is much faster.}
We visualize the average design time on the test set of all models in Figure~\ref{figure_analysis} (a). 
It shows our model is much more efficient than all other competitors. Inpainting and PROTSEED achieve the shortest inference time among all the baselines, but our model still outpaces them, being at least $17\times$ faster ($0.17$s compared to $3.00$s on $\beta$-lactamase).
The reason is the global attention sub-layer and neighborhood equivariant sub-layer update features for all residues in an one-shot manner, leading to a much more efficient inference process. 
The detailed design time of all models are provided in Appendix~\ref{appendix: infernce_time}.

\begin{table*}[!t]
\small
\centering
\begin{tabular}{lcccc}
\midrule
 Models & Top-5 ($\downarrow$) & Top-10 ($\downarrow$) & Top-30 ($\downarrow$) & Median ($\downarrow$) \\
\midrule
\cellcolor{myblue}\model & \cellcolor{myblue}$\textbf{-14.12$\pm$0.01}$ & \cellcolor{myblue}$\textbf{-13.85$\pm$0.10}$  & \cellcolor{myblue}$\textbf{-13.06$\pm$0.38}$ & \cellcolor{myblue}$\textbf{-10.74}$  \\
~--  w/o-gate & -12.74$\pm$0.40 & -12.50$\pm$0.37 & -12.07$\pm$0.41 & -9.63  \\
~--  w/o-kNN & -11.97$\pm$0.10 & -11.80$\pm$0.19 & -11.61$\pm$0.17 & -7.15 \\
~--  w/o-MFFN & -12.14$\pm$0.31 & -11.88$\pm$0.35 & -11.64$\pm$0.26 & -9.26 \\
~--  w/o-MCFFN  & -12.26$\pm$0.36 & -11.99$\pm$0.37 & -11.69$\pm$0.30 & -9.22 \\
~--  w-RandomInit & -13.52$\pm$0.30 & -12.90$\pm$0.66 & -12.11$\pm$0.68 & -9.09 \\
EGNN+ESM2 &-13.30$\pm$0.61 & -12.77$\pm$0.69  & -12.07$\pm$0.65 & -9.63 \\
\bottomrule
\end{tabular}
\caption{\model and variants on myoglobin design. Notice that computing message only on k-nearest neighbors rather than on the full residues contributes to significant improvement in \model. The unit system for top-5, top-10, top-30 mean and median binding affinity score is kcal/mol.}
\label{Tab: Ablation}
\end{table*}

\subsection{Ablation Study: Contribution of Model Components}
Table~\ref{Tab: Ablation} shows the results of ablation study on myoglobin.
The most substantial degradation in performance occurs when applying complete pairwise residue graph (\model-w/o-kNN) instead of k-nearest neighbors.
This outcome could be attributed to the excessively long protein sequences, where distant residues in 3D space may not have meaningful connections. Forcing these distant residues to exchange information might introduce noise into the model.
Replacing FFN with the self-attention mechanism during message update will also cause performance degradation~(\model-w/o-MFFN), while the additional removal of FFN in coordinate update~(w/o-MCFFN) does not lead to a significant performance difference.
It demonstrates that FFN in message update can better gather information by considering both semantic and structural similarities.
Leveraging the gating mechanism will slightly boost the model performance, and utilizing ESM2 initialization can further promote functional protein design.
% To further validate the effectiveness of gating mechanism, we conduct an additional test in a more challenging setting: $\beta$-lactamase with $30\%$ identity threshold.
% The results are reported in Appendix Table~\ref{Tab: Ablation_gating}. It shows that without gating mechanism, the model performance degrades a lot, demonstrating the robust performance of gating mechanism across various experimental scenarios.

%% file: 005analysis.tex
\label{Section:analysis_section}

\subsection{Effects of MSA Identity Threshold}
To gain insights into how the given meaningful fragments would influence the design quality, we provide the results on myoglobin under different MSA thresholds.
Figure~\ref{figure_analysis} (b) and (c) show a progressive decline in the quality of the designed sequences and structures as the MSA threshold increases. However, when the threshold remains below $30\%$, there are no substantial differences observed.
To ensure high design quality while providing minimum information, we set the threshold to $30\%$ for myoglobin. This choice aims to strike a balance between generating plausible proteins and fostering novelty.

\begin{figure}
\begin{minipage}[t]{0.33\linewidth}
\centering
\includegraphics[width=2.5cm]{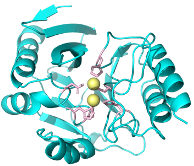}
\centerline{(a) B1 metallo-$\beta$-lactamase}
\end{minipage}%
\begin{minipage}[t]{0.33\linewidth}
\centering
\includegraphics[width=2.6cm]{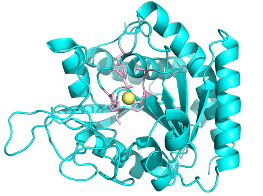}
\centerline{(b) B2 metallo-$\beta$-lactamase}
\end{minipage}%
\begin{minipage}[t]{0.33\linewidth}
\centering
\includegraphics[width=2.4cm]{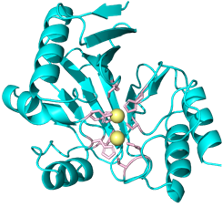}
\centerline{(c) B3 metallo-$\beta$-lactamase}
\end{minipage}
\vspace{-0.6em}
	\caption{Designed $\beta$-lactamases belonging to different subclasses: (a) B1, (b) B2 and (c) B3 metallo-$\beta$-lactamases.} 
 \label{figure_beta}
\end{figure}

\begin{figure}
\begin{minipage}[t]{0.33\linewidth}
\centering
\includegraphics[width=2.5cm]{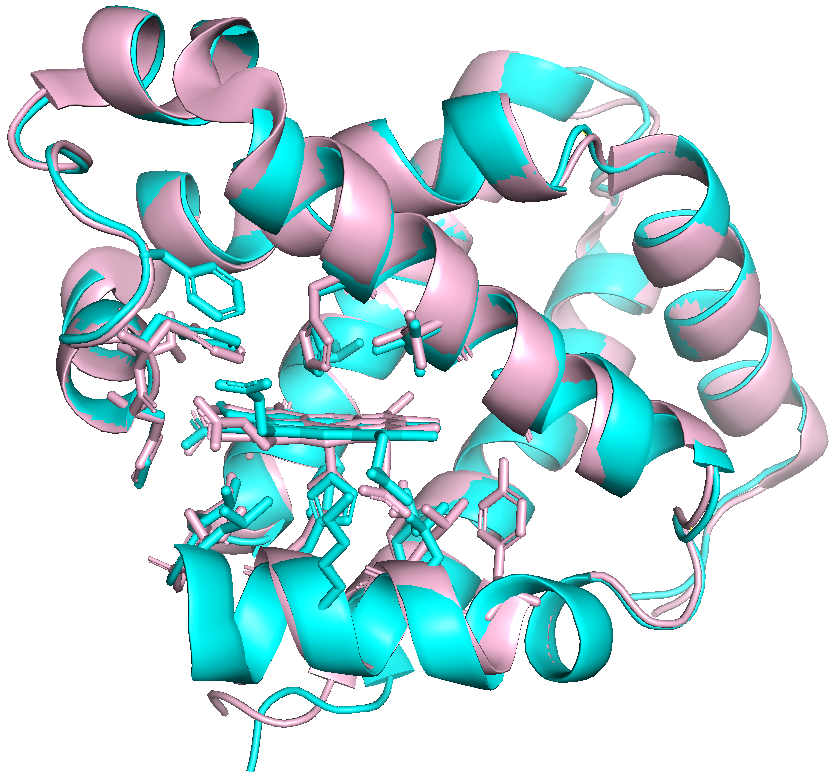}
\centerline{(a) binding at His-92}
\end{minipage}%
\begin{minipage}[t]{0.33\linewidth}
\centering
\includegraphics[width=2.5cm]{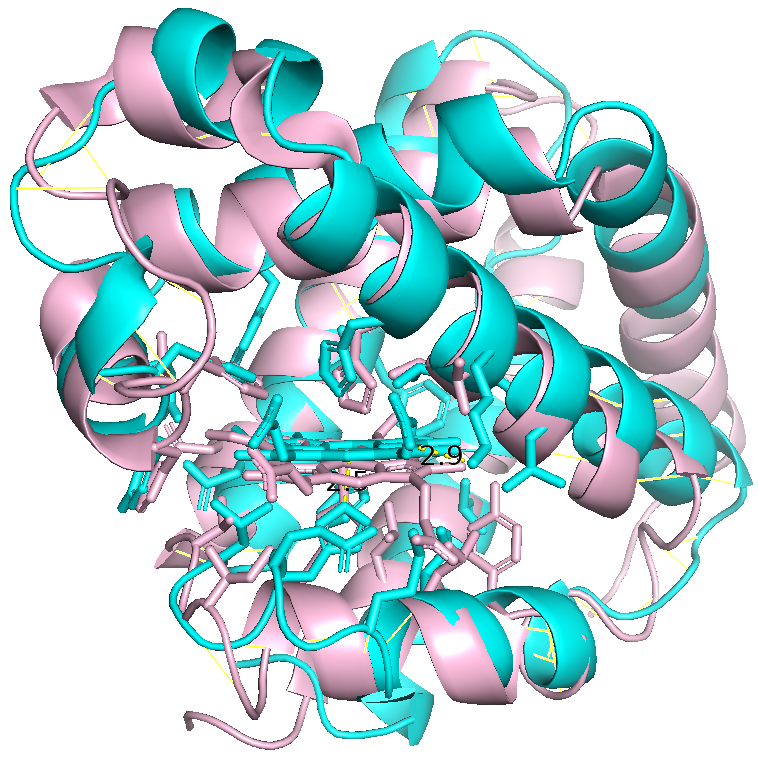}
\centerline{(b) Binding at His-89}
\end{minipage}%
\begin{minipage}[t]{0.33\linewidth}
\centering
\includegraphics[width=2.5cm]{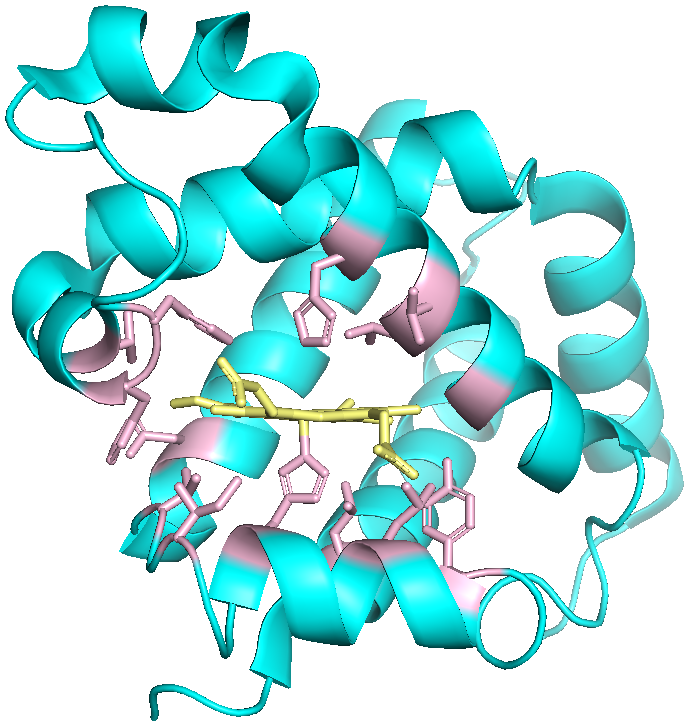}
\centerline{(c) myoglobin with PDB id=1SPG}
\end{minipage}
\vspace{-0.6em}
 \caption{Designed myoglobins binding heme ligand at (a) 92-Histidine and (b) 89-Histidine, and respectively
 have a low amino acid identity rate of $66.0\%$ and $26.7\%$ to the most similar one in Uniprot~((c) PDB id=1SPG) but also with a low RMSD distance of (a) $0.458${\small \AA} \space and (b) $3.943${\small \AA}.} 
 \label{figure_myoglobin}
\end{figure}

\subsection{Can \model designed proteins bind the corresponding metallocofactors?}
To further test if our designed proteins are potential to have the desired functions, we evaluate the basic properties of metalloproteins, i.e. if they can bind their corresponding metallocofactors.
Metalloproteins are required to contain metal ion cofactors to carry out their functions, and thus assessing the metallocofactor binding property is meaningful and important. 
Specifically, we first randomly select $20$ cases from the top-100 sequences according to their pLDDT scores.
We then employ AlphaFold2 for protein structure prediction, followed by inputting these structures into AlphaFill~\citep{hekkelman2023alphafill} to predict the associated ligands. 
If the predicted complex by AlphaFill includes the corresponding metal ion, we think the designed protein is highly potential to bind the metallocofactor.
Our results indicate that all designed myoglobins are predicted to have a heme ligand, while 18 $\beta$-lactamases are predicted to have zinc ion ligand. 
\textbf{It shows that our method can generate proteins that are highly potential to bind the corresponding metallocofactors.}

\textbf{We also find our model is capable of designing diverse proteins.}
We provide three designed $\beta$-lactamases in Figure~\ref{figure_beta} and find they have almost the same active site environments to those of B1, B2 and B3 metallo-$\beta$-lactamases reported in ~\citet{palzkill2013metallo}~(assessment process is provided in Appendix~\ref{active_site}).
It validates that our method can design proteins with diverse structures. 

\textbf{Our method is also able to design novel proteins.}
We blast the $20$ designed proteins in Uniprot and find two myoglobins that have low amino acid identity rates to the most similar one~(PDB id=1SPG, Figure~\ref{figure_myoglobin}(c)) in database, say $66.0\%$ and $26.7\%$ respectively. It validates that our model has the ability to design novel protein sequences.
Then we overlay their structures with 1SPG as shown in Figure~\ref{figure_myoglobin} (a) and (b), of which (a) has almost the same structure (RMSD 0.458{\small \AA}) and (b) has similar active site environments and is predicted to have heme ligand by AlphaFill but is slightly different with 1SPG~(RMSD 3.943{\small \AA}). It demonstrates the generated protein sequences by our model also have reasonable and novel structures.
We provide the sequences for all above cases and more designed samples in Appendix~\ref{appendix_cases}.

%% file: 007conclusion.tex
This paper proposes \model, a method to co-design protein sequence and backbone structure based on automatically detected meaningful fragments.
\model consists of stacked novel {\layer}s, each of which includes a global attention sub-layer and a neighborhood equivariant sub-layer to facilitate comprehensive information exchange. 
Experimental results show that our \model is able to design novel proteins with desired functions. 
One limitation of this work is, although our model demonstrats promising results, the designed proteins have not undergone wet-lab testing. 
Future work will involve the wet-lab testing to verify the biological functions of \model designed proteins.

%% file: 008appendix.tex
\section*{Appendix}
\section{Proof of Theorem 3.1 and Corollary 3.2}
\label{proof_section_3_5}

\subsection{Proof of Theorem 3.1}
\label{Appendix: theorem 3.1}
In this section we prove that our proposed \layer is translation equivariant on $\boldsymbol{x}$ for any translation vector $\boldsymbol{t}\in \mathbb{R}^3$ and rotation equivariant for any $R$ from SO(3) group. More formally, we will prove the \layer satisfies:
\begin{equation}
\small 
\boldsymbol{H}^{l+1}, R\boldsymbol{x}^{l+1}+\boldsymbol{t} = \mathrm{\layer}(\boldsymbol{H}^{l}, R\boldsymbol{x}^{l}+\boldsymbol{t})
\end{equation}

When $l=0$:\\
For message update, the distance between two residues is invariant as $d_{ij}^{l} = ||R\boldsymbol{x}_i^{l}+\boldsymbol{t}-(R\boldsymbol{x}^{l}_j+\boldsymbol{t})||^2=(\boldsymbol{x}_i^{l}-\boldsymbol{x}^{l}_j)^TR^TR(\boldsymbol{x}_i^{l}-\boldsymbol{x}^{l}_j)=||\boldsymbol{x}_i^{l}-\boldsymbol{x}^{l}_j||^2$. 
$\boldsymbol{h}^{l}_i$ is the embedding of residue or [mask] token, and thus it is always invariant.
Therefore, $\boldsymbol{h}^{l+0.5}_i$ is also invariant.
Since $\boldsymbol{h}^{l+0.5}_i$, $\boldsymbol{h}^{l+0.5}_j$ and $d_{ij}^{l}$ are all invariant, and thus $\boldsymbol{m}_{ij}^{l+1}$ is also invariant to translation $\boldsymbol{t}$ and rotation $R$ on $\boldsymbol{x}$.

For coordinate update, updated $\boldsymbol{x}^{l+1}$ is equivariant to the translation $\boldsymbol{t}$ and rotation $R$ on input $\boldsymbol{x}^l$:
\begin{equation}
\begin{split}
&(R\boldsymbol{x}^{l}_i+\boldsymbol{t})+\sum_{j\in \mathrm{Neighbor(i)}}(R\boldsymbol{x}^{l}_i+\boldsymbol{t}-(R\boldsymbol{x}^{l}_j+\boldsymbol{t}))\cdot \mathrm{FFN}(\boldsymbol{m}_{ij}^{l+1})\\
&=R(\boldsymbol{x}^{l}_i+\sum_{j\in \mathrm{Neighbor(i)}}(\boldsymbol{x}^{l}_i-\boldsymbol{x}^{l}_j)\cdot \mathrm{FFN}(\boldsymbol{m}_{ij}^{l+1})) + \boldsymbol{t} \\
&=R\boldsymbol{x}^{l+1}_i + \boldsymbol{t}
\end{split}
\end{equation}

For residue update, $\boldsymbol{m}_{ij}^{l+1}$ and $\boldsymbol{h}_i^{l+0.5}$ is invariant to translation $\boldsymbol{t}$ and rotation $R$ on $\boldsymbol{x}$, so $\boldsymbol{h}_i^{l+1}$ is also invariant to translation $\boldsymbol{t}$ and rotation $R$ on $\boldsymbol{x}$.

When $l>=1$:\\
We have proved that when $l=0$, $\boldsymbol{h}^1_i$ is invariant to rotation and translation on $\boldsymbol{x}$. Taking $\boldsymbol{H}^1$ as the second layer input and following the above process, we can prove $\boldsymbol{h}^2_i$ is also invariant. Repeating this process from $l=1$ to $L-1$, we can get the same conclusion.

Combining the above two scenarios together, we have $\boldsymbol{H}^{l+1}, R\boldsymbol{x}^{l+1}+\boldsymbol{t} = \mathrm{\layer}(\boldsymbol{H}^{l}, R\boldsymbol{x}^{l}+\boldsymbol{t})$ for $l=0$ to $L-1$. Therefore, our proposed \layer is $SE(3)$-equivariant.

\subsection{Proof of Corollary 3.2}
\label{proof: corollary_3_2}

We provide the proof of corollary 3.2 as follows:
\begin{proof}\renewcommand{\qedsymbol}{}
\begin{equation}
\small
\begin{split}
\mathrm{\model}(\boldsymbol{H}^{0}, R\boldsymbol{x}^{0}+\boldsymbol{t}) &= \mathrm{\layer}^{L-1}\circ \mathrm{\layer}^{L-2}\circ \cdot \cdot \cdot \circ \mathrm{\layer}^{0}(\boldsymbol{H}^{0}, R\boldsymbol{x}^{0}+\boldsymbol{t}) ) \\
&= \mathrm{\layer}^{L-1}\circ \mathrm{\layer}^{L-2}\circ \cdot \cdot \cdot \circ \mathrm{\layer}^{1}(\boldsymbol{H}^{1}, R\boldsymbol{x}^{1}+\boldsymbol{t})) \\
&=...\\
&=\mathrm{\layer}^{L-1}(\boldsymbol{H}^{L-1}, R\boldsymbol{x}^{L-1}+\boldsymbol{t})) = \boldsymbol{H}^{L}, R\boldsymbol{x}^{L}+\boldsymbol{t}
\end{split}
\end{equation} 
\end{proof}

% \begin{figure*}[h]
%   \centering
% \includegraphics[width=12.0cm]{chirality.pdf}
%   \vspace{-0.6em}
%   \caption{An example to illustrate that our model does not satisfy reflection equivariance. Here, left side sequence is Ala-Val which is made by natural amino acids. On the right side, the sequence is also Ala-Val. The reflection will not affect the sequence. However, the conformation of the amino acid did change from L-amino acid to D-amino acid.}
%   \label{figure_glat}
% \end{figure*}

\section{Additional Experimental Details}

\subsection{Significance of Metalloproteins Studied Herein}
\label{reason_for_datasets}
Metalloproteins comprise almost 50\% of all the naturally occuring proteins. 
The Zn-dependent $\beta$-lactamases and heme-dependent myoglobins studied herein represent biologically signicant metalloprotein examples.
In particular, $\beta$-lactamases are enzymes produced by microorganisms to break down $\beta$-lactam antibiotics, conferring antibiotic resistance~\citep{gupta2008metallo}. Thus, the study and design of $\beta$-lactamases hold relevance to public health and play a critical role in the development of new antibiotics.
On the other hand, myoglobin is a heme-containing protein involved in oxygen storage and transport in muscle tissue~\citep{springer1994mechanisms}, highlighting their biological significance.
Note that the structure of hemoglobin closely resembles that of myoglobin~\citep{antonini1965interrelationship}, so we merge these two metalloproteins and denote the merged dataset simply as myoglobin.

\subsection{Protein Data Statistics}
\label{appendix:data_statistics}
Detailed data statistics for $\beta$-lactamase and myoglobin are reported in Table~\ref{Tab: data_statistics}.
We will release the two cleaned metalloprotein datasets in the near future.

\begin{table}[!t]
\small
\begin{center}
\begin{tabular}{lccc}
\midrule
Protein  & PDB & Metal Binding & Length Filtering  \\
\midrule
$\beta$-lactamase & $171,484$ & $7,802$ & $5,427$\\
myoglobin & $14,573$ & $3,381$ & $3,381$ \\
\bottomrule
\end{tabular}
\end{center}
\caption{Detailed data statistics of the two metalloproteins.}
\label{Tab: data_statistics}
\end{table}

\begin{table}[!t]
\small
\begin{center}
\begin{tabular}{lcc}
\midrule
Models & $\beta$-lactamase & myoglobin \\
\midrule
Haluccination & $612.60$  & $564.60$ \\ 
Inpainting & $3.00$ & $3.07$ \\
SMCDiff+ProteinMPNN & $237.67$ & $105.60$\\
PROTSEED & $4.87$ & $3.05$ \\
FrameDiff+ProteinMPNN & $124.95$ & $68.65$ \\
RFDiffusion+ProteinMPNN & $26.20$ & $25.60$\\
\cellcolor{myblue}\model &\cellcolor{myblue}$\textbf{0.17}$ & \cellcolor{myblue}$\textbf{0.07}$ \\
\bottomrule
\end{tabular}
\end{center}
\caption{Average inference time (s) for the two metalloproteins on test set.}
\label{Tab: infer_time}
\end{table}

% \subsection{More Implementation Details}
% \label{appendix_implementation_details}
% We apply Adam~\cite{kingma2014adam} as the optimizer with a linear warm-up over the first $4,000$ steps and linear decay for later steps.

% \begin{table*}[!t]
% \small
% \centering
% \begin{tabular}{lccccc}
% \midrule
% Models & AAR (\%,$\uparrow$) & RMSD (\AA,$\downarrow$) & pLDDT ($\uparrow$) & TM-score ($\uparrow$) & Consistency (\AA,$\downarrow$)   \\
% \midrule
% \cellcolor{myblue}\model & \cellcolor{myblue}$\textbf{66.89}$ & \cellcolor{myblue}$\textbf{2.8725}$ & \cellcolor{myblue}$63.5121$   &\cellcolor{myblue}$\textbf{0.4341}$ &\cellcolor{myblue}$\textbf{9.4449}$  \\
% ~--  w/o-gate & $60.98$ & $2.8907$ & $\textbf{63.6199}$ & $0.3753$ & $10.0390$ \\
% \bottomrule
% \end{tabular}
% \vspace{-0.5em}
% \caption{Ablation study for gating mechanism on $\beta$-lactamase with $30\%$ identity threshold.}
% \label{Tab: Ablation_gating}
% \end{table*}

\subsection{Active Site Environment Assessment}
\label{active_site}
We outline the procedures for verifying whether the designed metalloproteins share similar active site environments with their natural counterparts.
For $\beta$-lactamase, we detect the residues that directly contact with zinc ion. If they show similar chemistry properties with natural $\beta$-lactamases, we think they can highly potentially bind to the corresponding metallocofactors.
For myoglobin, we calculate the distance between axial histidine ligand to Fe ion.
Besides, we also detect the presence of distal histidine within the active sites, which plays an important role in the natural function of myoglobin, i.e., oxygen molecule binding. 
If the distance is between 2.0\AA \space and 2.5\AA, and the distal histidine exists, we think the designed myoglobins are highly potentially to bind heme.

\section{Additional Experiments}

\subsection{Average Inference Time}
\label{appendix: infernce_time}
We provide detailed inference time of all models on the two datasets in Table~\ref{Tab: infer_time}. It shows our \model has much faster inference speed than all other competitors.
The reason is that our proposed \layer update messages, residue coordinates and residue representations of all residues in an one-shot manner, leading to a much more efficient design process.

% \subsection{Attention Visualization}
% We randomly pick one sentence and provide the attention matrix visualization from the last layer in Figure~\ref{figure}.

% \begin{figure}
% \begin{minipage}[t]{0.5\linewidth}
% \centering
% \includegraphics[width=5cm]{attn.png}
% \centerline{(a) attention matrix}
% \end{minipage}%
% \begin{minipage}[t]{0.5\linewidth}
% \centering
% \includegraphics[width=5cm]{co-occur.png}
% \centerline{(b) amino acid co-occurrence}
% \end{minipage}
% 	\caption{Visualization of (a) attention matrix. (b) pairwise amino acid co-occurrence frequency.} 
%  \label{figure_analysis}
% \end{figure}

% \subsection{Additional Ablation Study on Gating Mechanism}
% \label{appendix: ablation_gating}
% To further validate the effectiveness of gating mechanism, we conduct an ablation study in a more challenging setting: $\beta$-lactamase with a $30\%$ identity threshold.
% $\beta$-lactamase, characterized by diverse structures across its three sub-classes and sequences exceeding 1000 residues in length, presents a much harder testing ground.
% Table~\ref{Tab: Ablation_gating} reports the results and we can see that without gating mechanism, the model performance degrades a lot in a more challenging setting. It validates the stable expressiveness of gating mechanism across general experimental settings.

\begin{table*}[!t]
\tiny
\centering
\begin{tabular}{ll}
\midrule
Case & Protein Sequence \\
\midrule
Figure~\ref{figure_beta} (a) & EHSVVEISDDISITQLSDKVYTYVSLAEIEGWGMVPSNGMIVINNHQAALLDTPINDAQTEMLVNWVTDSLHAKVTTFIPNHWH\\
& GDCIGGLGYLQRKGVQSYANQMTIDLAKEKGLPVPEHGFTDSLTVSLDGMPLQCYYLGGGHATDNIVVWLPTENILFGGCMLK\\
& DNQTTSIGNISDADVTAWPKTLDKVKAKFPSARYVVPGHGNYGGTELIEHTKQIVNQYIESTS\\
Figure~\ref{figure_beta} (b) & GHSYEKYNNWETIEAWTKQVTSENPDLISRTAIGTTFLGNNIYLLKVGKPGPNKPAIFMDCGIHAREWISHAFCQWFVREAVLTY\\
&GYESHMTEFLNKLDFYVLPVLNIDGYIYTWTKNRMWRKTRSTNAGTTCIGTDPNRNFDAGWCTTGASTDPCDETYCGSAAESE\\
&KETKALADFIRNNLSSIKAYLSIHSYSQHIVYPYSYDYKLPENNAELNNLAKAAVKELATLYGTKYTYGPGATTLYLAPGGGDDW\\
&AYDQGIKYSFTFELRDKGRYGFILPESQIQATCEETMLAIKYVTNYVLGHLY \\
Figure~\ref{figure_beta} (c) &TVIKNETGTISISQLNKNVWVHTELGVPSNGLVLNTSKGLVLVDSSWDDKLTKELIEMVEKKFQKRVTDVIITHAHADHIGGIKTL\\
&KERGIKAHSTALTAELAKKNGYEEPLGDLQTVTNLKFGNMKVETFYPGKGHTEDNIVVWLPQYNILVGGSLVKSTSAKDLGNVV\\
&ADAYNEWSTSIENVLKRYRNINAVVPGHGEVGDKGLLLHTLDLLK \\
\midrule
Figure~\ref{figure_myoglobin} (a) &VDWTDAERAAITDLWAKVDVEDVGAQALARLLVVYPWTQRYFGGFGNISSASAILGNAKVAAHGKTVLTGLDRAIAHMDDIAG\\
&AFTQLSVKHSEKLHVDPDNFKVVGDLLTIVLAAVLGADFTPEVKAAWQKFLAVIVSALSRRYH \\
Figure~\ref{figure_myoglobin} (b) & GFKQDIATIRGDLRTYAQDIFLAFLNKYPDERRYFKNYVGKSDQELKSMAKFGDHTEKVFNLMMEVADRATDCVPLASDANTLV\\
& QMKQHSSLTTGNFEKLFVALVEYMRASGQSFDSQSWDRFGKNLVSALSSAGMK \\
\bottomrule
\end{tabular}
\vspace{-0.5em}
\caption{Protein sequences for cases in Figure~\ref{figure_beta} and \ref{figure_myoglobin}.}
\label{Tab: protein_seq}
\end{table*}

\section{Designed Protein Display}
\label{appendix_cases}
\subsection{Novel and Diverse Protein Sequences}
We report the protein sequences shown in Figure~\ref{figure_beta} and \ref{figure_myoglobin} in Table~\ref{Tab: protein_seq}.

\subsection{More Designed Cases}
\begin{figure}[htbp]
\centering
\subfigure[2ANP.A]{
\begin{minipage}[t]{0.33\linewidth}
\centering
\includegraphics[width=3.5cm]{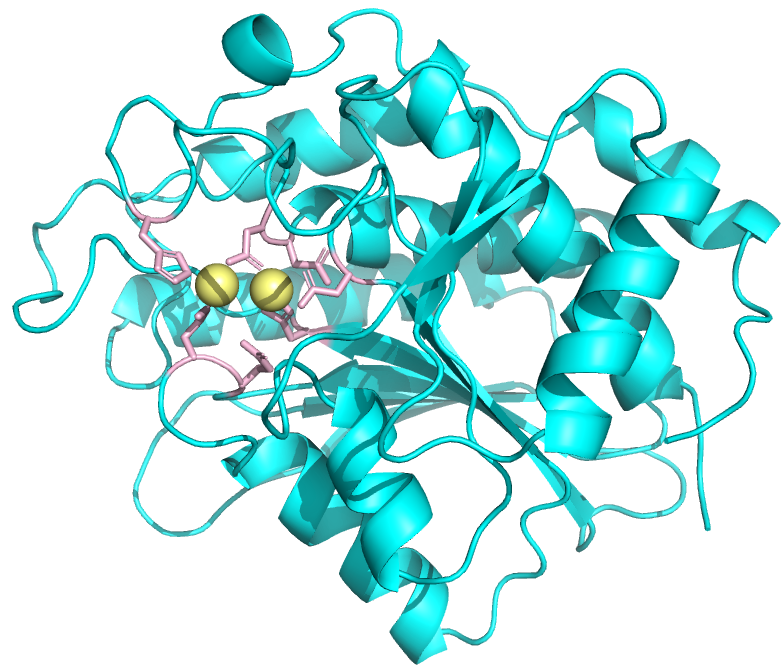}
%\caption{fig1}
\end{minipage}%
}%
\subfigure[3WUB.A]{
\begin{minipage}[t]{0.33\linewidth}
\centering
\includegraphics[width=3.5cm]{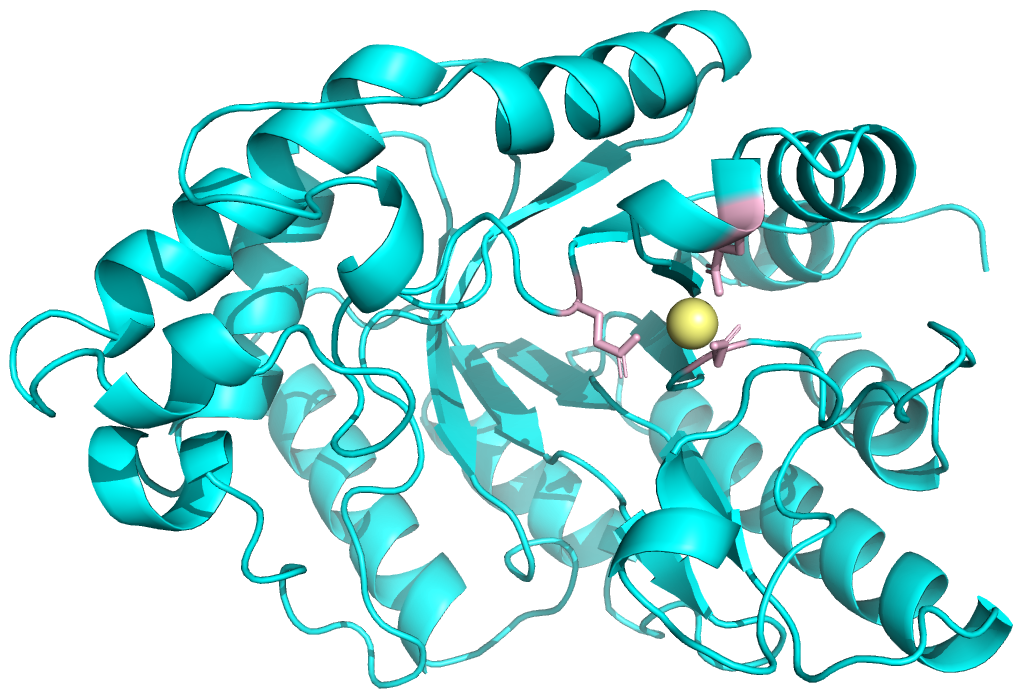}
%\caption{fig2}
\end{minipage}%
}%
\subfigure[3TLI.A]{
\begin{minipage}[t]{0.33\linewidth}
\centering
\includegraphics[width=3.5cm]{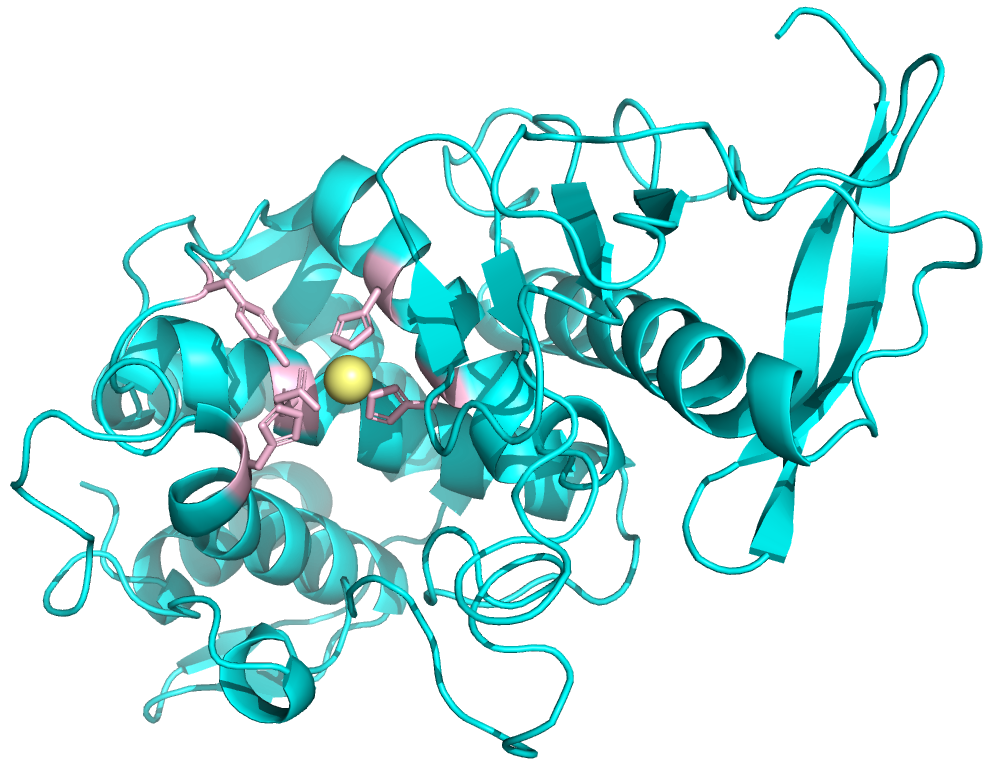}
%\caption{fig2}
\end{minipage}%
}%
\quad                
\subfigure[1TKF.A]{
\begin{minipage}[t]{0.33\linewidth}
\centering
\includegraphics[width=3.5cm]{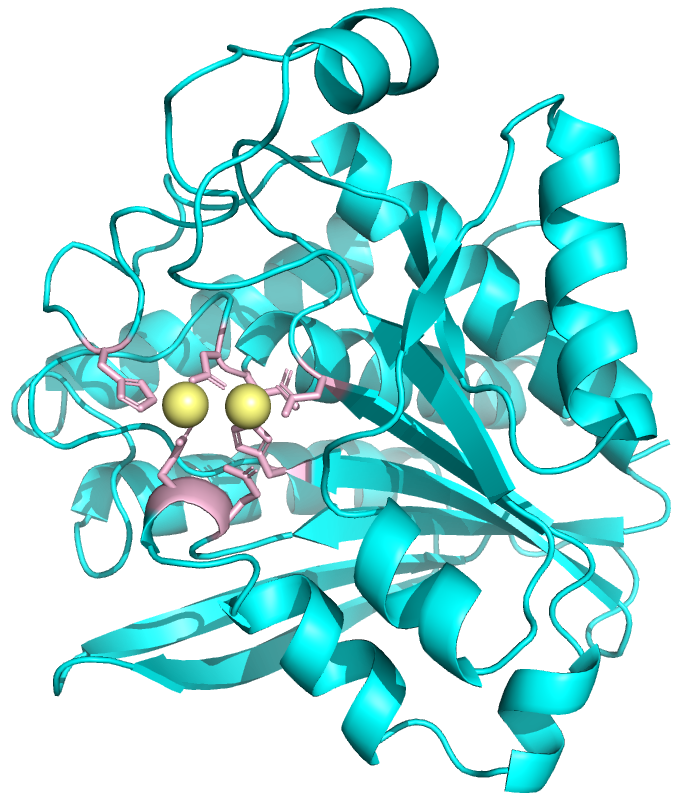}
%\caption{fig2}
\end{minipage}
}%
\subfigure[3FX6.A]{
\begin{minipage}[t]{0.33\linewidth}
\centering
\includegraphics[width=3.5cm]{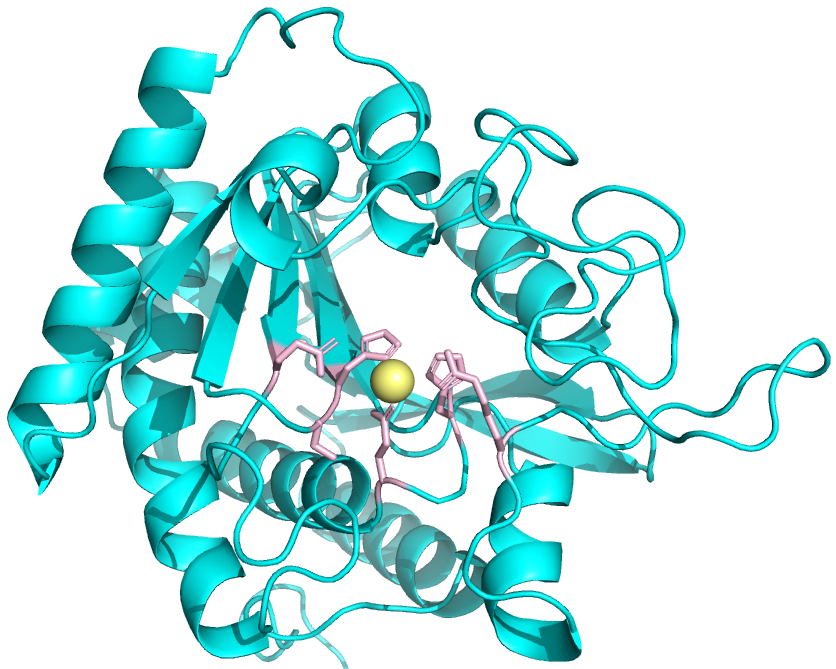}
%\caption{fig2}
\end{minipage}
}%
\subfigure[5YPM.A]{
\begin{minipage}[t]{0.33\linewidth}
\centering
\includegraphics[width=3.5cm]{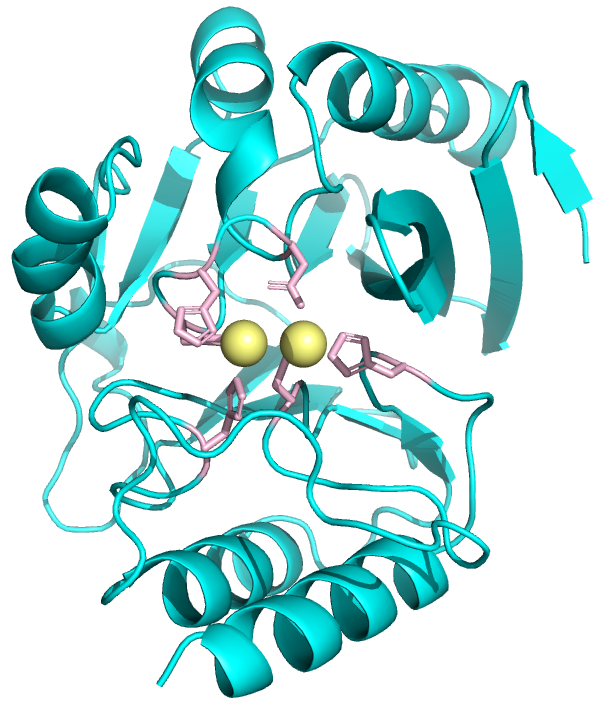}
%\caption{fig2}
\end{minipage}%
}%
\quad                 
\subfigure[1IY7.A]{
\begin{minipage}[t]{0.33\linewidth}
\centering
\includegraphics[width=3.5cm]{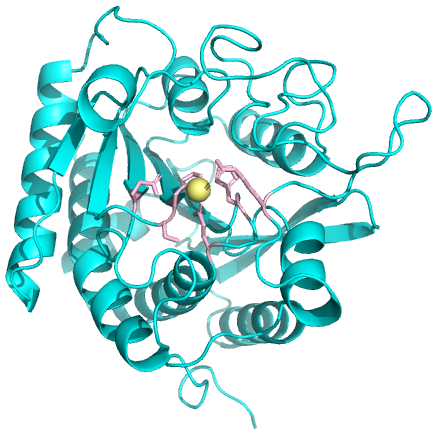}
%\caption{fig2}
\end{minipage}
}%
\subfigure[1A8T.A]{
\begin{minipage}[t]{0.33\linewidth}
\centering
\includegraphics[width=3.5cm]{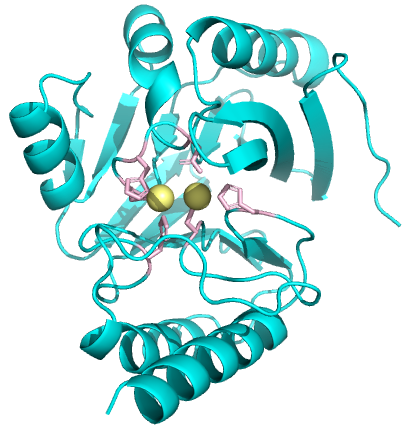}
%\caption{fig2}
\end{minipage}
}%
\subfigure[4Q6P.A]{
\begin{minipage}[t]{0.33\linewidth}
\centering
\includegraphics[width=3.5cm]{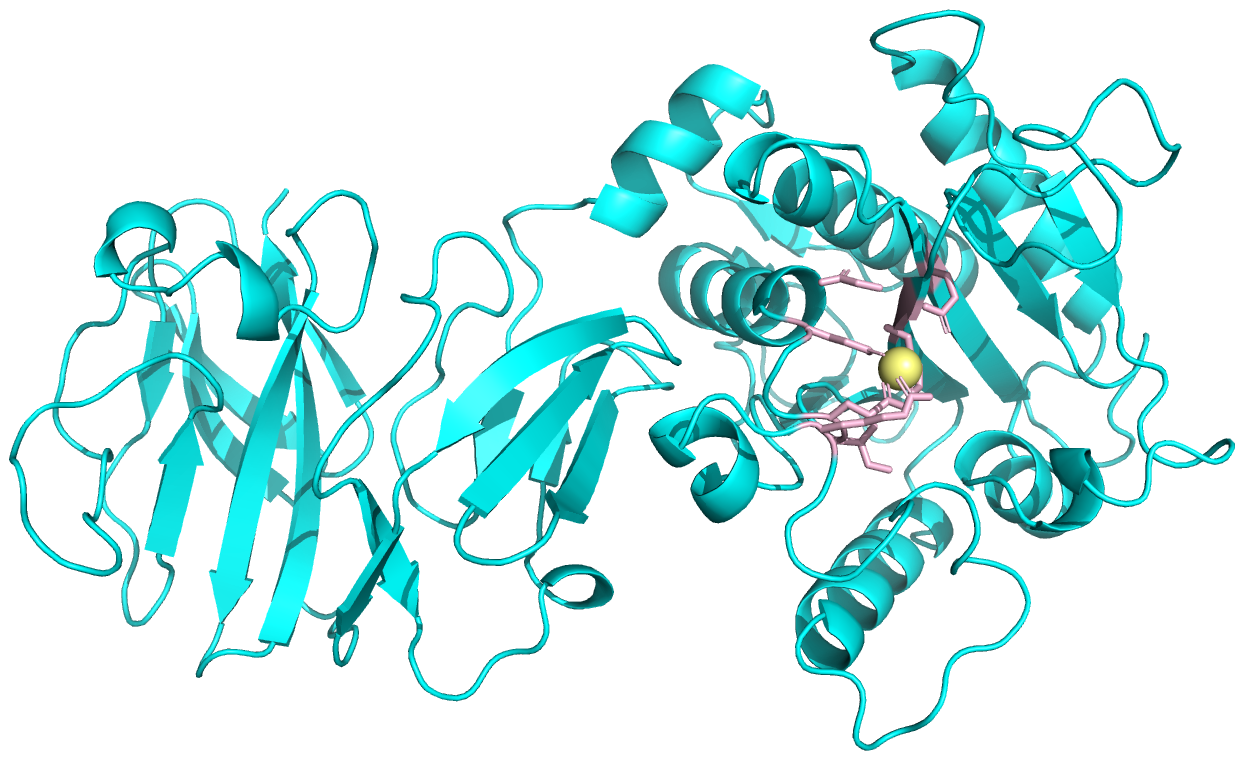}
%\caption{fig2}
\end{minipage}%
}%
\centering
\caption{$\beta$-lactamases with motifs identical to those found in the target proteins in the PDB.}
\label{Fig:appendix_case_beta}
\end{figure}

\begin{figure}[htbp]
\centering
\subfigure[1S5Y.B]{
\begin{minipage}[t]{0.33\linewidth}
\centering
\includegraphics[width=3.5cm]{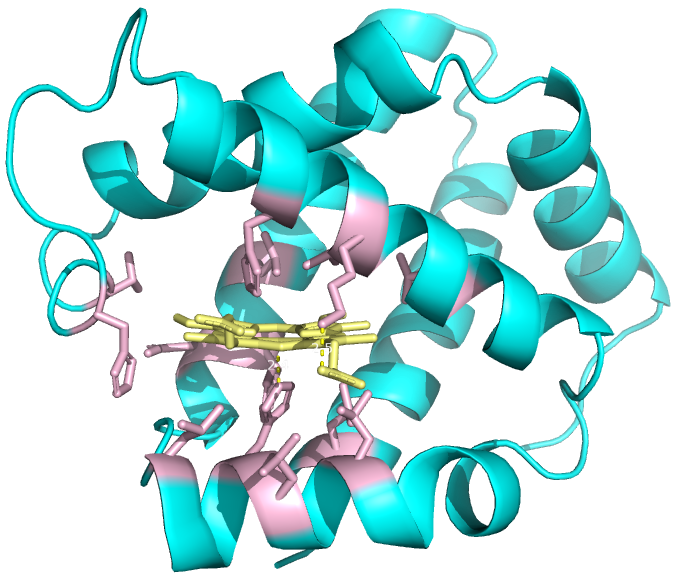}
%\caption{fig1}
\end{minipage}%
}%
\subfigure[1Y5F.C]{
\begin{minipage}[t]{0.33\linewidth}
\centering
\includegraphics[width=3.5cm]{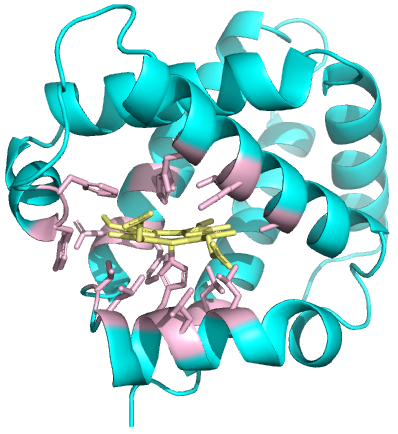}
%\caption{fig2}
\end{minipage}%
}%
\subfigure[1JEB.C]{
\begin{minipage}[t]{0.33\linewidth}
\centering
\includegraphics[width=3.5cm]{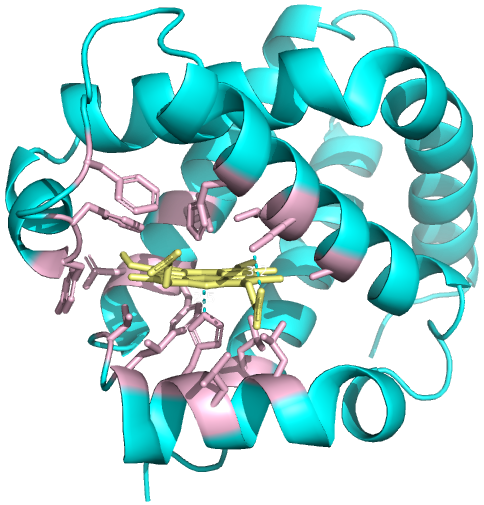}
%\caption{fig2}
\end{minipage}%
}%
\quad                 
\subfigure[3G53.A]{
\begin{minipage}[t]{0.33\linewidth}
\centering
\includegraphics[width=3.5cm]{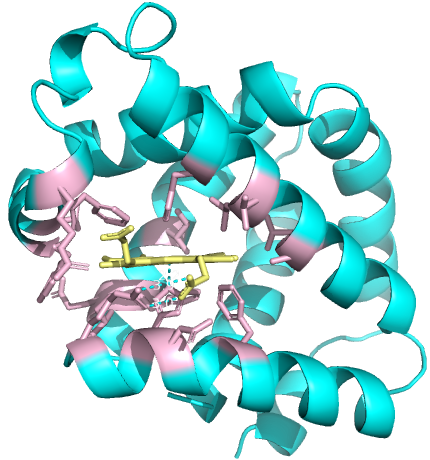}
%\caption{fig2}
\end{minipage}
}%
\subfigure[7AEU.B]{
\begin{minipage}[t]{0.33\linewidth}
\centering
\includegraphics[width=3.5cm]{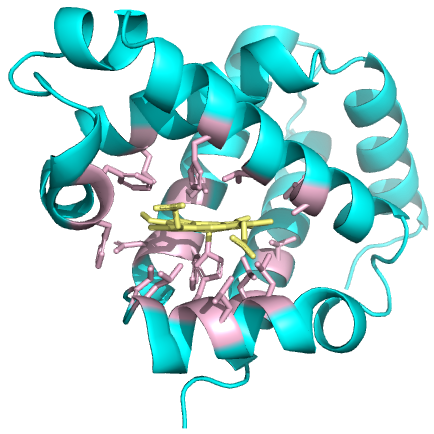}
%\caption{fig2}
\end{minipage}
}%
\subfigure[2ZLU.B]{
\begin{minipage}[t]{0.33\linewidth}
\centering
\includegraphics[width=3.5cm]{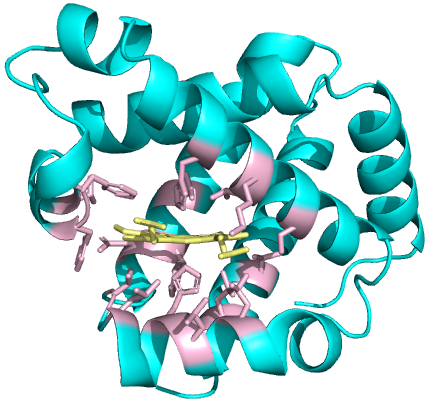}
%\caption{fig2}
\end{minipage}%
}%
\quad                 
\subfigure[5E83.C]{
\begin{minipage}[t]{0.33\linewidth}
\centering
\includegraphics[width=3.5cm]{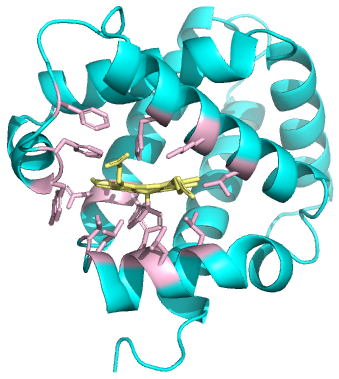}
%\caption{fig2}
\end{minipage}
}%
\subfigure[1X9F.B]{
\begin{minipage}[t]{0.33\linewidth}
\centering
\includegraphics[width=3.5cm]{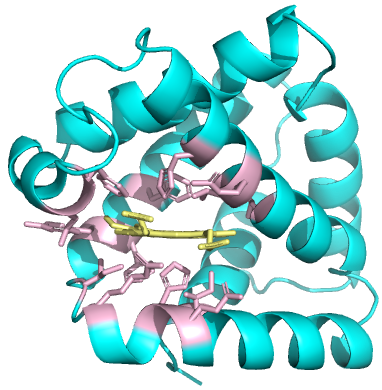}
%\caption{fig2}
\end{minipage}
}%
\subfigure[4NI1.B]{
\begin{minipage}[t]{0.33\linewidth}
\centering
\includegraphics[width=3.5cm]{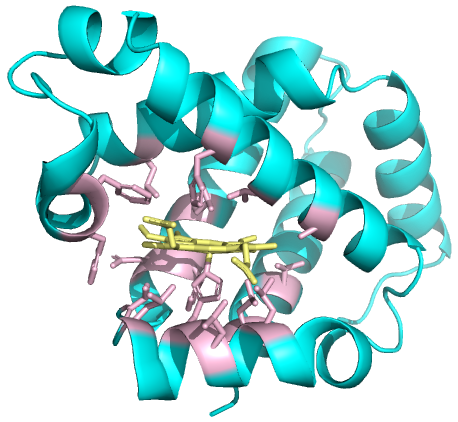}
%\caption{fig2}
\end{minipage}%
}%
\centering
\caption{myoglobins with motifs identical to those found in the target proteins in the PDB.}
\label{Fig:appendix_case_myoglobin}
\end{figure}

Figure~\ref{Fig:appendix_case_beta} and \ref{Fig:appendix_case_myoglobin} respectively illustrate more designed $\beta$-lactamases and myoglobins. 
It shows all proteins exhibit active site environments reminiscent of those of natural ones and can bind to corresponding metallofactors, i.e., $\beta$-lactamases bind zinc ion and myoglobins bind heme, demonstrating their excellent potential to be biologically functional.
Besides, these protein sequences are not exactly the same as those in PDB, demonstrating our \model has the ability to design novel proteins with desired functions.

%% file: iclr2024_conference.bbl
\begin{thebibliography}{49}
\providecommand{\natexlab}[1]{#1}
\providecommand{\url}[1]{\texttt{#1}}
\expandafter\ifx\csname urlstyle\endcsname\relax
  \providecommand{\doi}[1]{doi: #1}\else
  \providecommand{\doi}{doi: \begingroup \urlstyle{rm}\Url}\fi

\bibitem[Anand \& Achim(2022)Anand and Achim]{anand2022protein}
Namrata Anand and Tudor Achim.
\newblock Protein structure and sequence generation with equivariant denoising diffusion probabilistic models.
\newblock \emph{arXiv preprint arXiv:2205.15019}, 2022.

\bibitem[Anderson et~al.(2011)Anderson, Strope, and Moriyama]{anderson2011suitemsa}
Catherine~L Anderson, Cory~L Strope, and Etsuko~N Moriyama.
\newblock Suitemsa: visual tools for multiple sequence alignment comparison and molecular sequence simulation.
\newblock \emph{BMC bioinformatics}, 12\penalty0 (1):\penalty0 1--14, 2011.

\bibitem[Andrieu et~al.(2003)Andrieu, De~Freitas, Doucet, and Jordan]{andrieu2003introduction}
Christophe Andrieu, Nando De~Freitas, Arnaud Doucet, and Michael~I Jordan.
\newblock An introduction to mcmc for machine learning.
\newblock \emph{Machine learning}, 50:\penalty0 5--43, 2003.

\bibitem[Angermueller et~al.(2019)Angermueller, Dohan, Belanger, Deshpande, Murphy, and Colwell]{angermueller2019model}
Christof Angermueller, David Dohan, David Belanger, Ramya Deshpande, Kevin Murphy, and Lucy Colwell.
\newblock Model-based reinforcement learning for biological sequence design.
\newblock In \emph{International conference on learning representations}, 2019.

\bibitem[Anishchenko et~al.(2021)Anishchenko, Pellock, Chidyausiku, Ramelot, Ovchinnikov, Hao, Bafna, Norn, Kang, Bera, et~al.]{anishchenko2021novo}
Ivan Anishchenko, Samuel~J Pellock, Tamuka~M Chidyausiku, Theresa~A Ramelot, Sergey Ovchinnikov, Jingzhou Hao, Khushboo Bafna, Christoffer Norn, Alex Kang, Asim~K Bera, et~al.
\newblock De novo protein design by deep network hallucination.
\newblock \emph{Nature}, 600\penalty0 (7889):\penalty0 547--552, 2021.

\bibitem[Antonini(1965)]{antonini1965interrelationship}
Eraldo Antonini.
\newblock Interrelationship between structure and function in hemoglobin and myoglobin.
\newblock \emph{Physiological reviews}, 45\penalty0 (1):\penalty0 123--170, 1965.

\bibitem[Arnold(1998)]{arnold1998design}
Frances~H Arnold.
\newblock Design by directed evolution.
\newblock \emph{Accounts of chemical research}, 31\penalty0 (3):\penalty0 125--131, 1998.

\bibitem[Arnold(2018)]{arnold2018directed}
Frances~H Arnold.
\newblock Directed evolution: bringing new chemistry to life.
\newblock \emph{Angewandte Chemie International Edition}, 57\penalty0 (16):\penalty0 4143--4148, 2018.

\bibitem[Baek et~al.(2021)Baek, DiMaio, Anishchenko, Dauparas, Ovchinnikov, Lee, Wang, Cong, Kinch, Schaeffer, et~al.]{baek2021accurate}
Minkyung Baek, Frank DiMaio, Ivan Anishchenko, Justas Dauparas, Sergey Ovchinnikov, Gyu~Rie Lee, Jue Wang, Qian Cong, Lisa~N Kinch, R~Dustin Schaeffer, et~al.
\newblock Accurate prediction of protein structures and interactions using a three-track neural network.
\newblock \emph{Science}, 373\penalty0 (6557):\penalty0 871--876, 2021.

\bibitem[Belanger et~al.(2019)Belanger, Vora, Mariet, Deshpande, Dohan, Angermueller, Murphy, Chapelle, and Colwell]{belanger2019biological}
David Belanger, Suhani Vora, Zelda Mariet, Ramya Deshpande, David Dohan, Christof Angermueller, Kevin Murphy, Olivier Chapelle, and Lucy Colwell.
\newblock Biological sequences design using batched bayesian optimization.
\newblock 2019.

\bibitem[Brookes et~al.(2019)Brookes, Park, and Listgarten]{brookes2019conditioning}
David Brookes, Hahnbeom Park, and Jennifer Listgarten.
\newblock Conditioning by adaptive sampling for robust design.
\newblock In \emph{International conference on machine learning}, pp.\  773--782. PMLR, 2019.

\bibitem[Brookes \& Listgarten(2018)Brookes and Listgarten]{brookes2018design}
David~H Brookes and Jennifer Listgarten.
\newblock Design by adaptive sampling.
\newblock \emph{arXiv preprint arXiv:1810.03714}, 2018.

\bibitem[Dalby(2011)]{dalby2011strategy}
Paul~A Dalby.
\newblock Strategy and success for the directed evolution of enzymes.
\newblock \emph{Current opinion in structural biology}, 21\penalty0 (4):\penalty0 473--480, 2011.

\bibitem[Das et~al.(2021)Das, Sercu, Wadhawan, Padhi, Gehrmann, Cipcigan, Chenthamarakshan, Strobelt, Dos~Santos, Chen, et~al.]{das2021accelerated}
Payel Das, Tom Sercu, Kahini Wadhawan, Inkit Padhi, Sebastian Gehrmann, Flaviu Cipcigan, Vijil Chenthamarakshan, Hendrik Strobelt, Cicero Dos~Santos, Pin-Yu Chen, et~al.
\newblock Accelerated antimicrobial discovery via deep generative models and molecular dynamics simulations.
\newblock \emph{Nature Biomedical Engineering}, 5\penalty0 (6):\penalty0 613--623, 2021.

\bibitem[Dauparas et~al.(2022)Dauparas, Anishchenko, Bennett, Bai, Ragotte, Milles, Wicky, Courbet, de~Haas, Bethel, et~al.]{dauparas2022robust}
Justas Dauparas, Ivan Anishchenko, Nathaniel Bennett, Hua Bai, Robert~J Ragotte, Lukas~F Milles, Basile~IM Wicky, Alexis Courbet, Rob~J de~Haas, Neville Bethel, et~al.
\newblock Robust deep learning--based protein sequence design using proteinmpnn.
\newblock \emph{Science}, 378\penalty0 (6615):\penalty0 49--56, 2022.

\bibitem[Ding et~al.(2022)Ding, Nakai, and Gong]{ding2022protein}
Wenze Ding, Kenta Nakai, and Haipeng Gong.
\newblock Protein design via deep learning.
\newblock \emph{Briefings in bioinformatics}, 23\penalty0 (3):\penalty0 bbac102, 2022.

\bibitem[Fleishman et~al.(2011)Fleishman, Leaver-Fay, Corn, Strauch, Khare, Koga, Ashworth, Murphy, Richter, Lemmon, et~al.]{fleishman2011rosettascripts}
Sarel~J Fleishman, Andrew Leaver-Fay, Jacob~E Corn, Eva-Maria Strauch, Sagar~D Khare, Nobuyasu Koga, Justin Ashworth, Paul Murphy, Florian Richter, Gordon Lemmon, et~al.
\newblock Rosettascripts: a scripting language interface to the rosetta macromolecular modeling suite.
\newblock \emph{PloS one}, 6\penalty0 (6):\penalty0 e20161, 2011.

\bibitem[Gainza et~al.(2023)Gainza, Wehrle, Van Hall-Beauvais, Marchand, Scheck, Harteveld, Buckley, Ni, Tan, Sverrisson, et~al.]{gainza2023novo}
Pablo Gainza, Sarah Wehrle, Alexandra Van Hall-Beauvais, Anthony Marchand, Andreas Scheck, Zander Harteveld, Stephen Buckley, Dongchun Ni, Shuguang Tan, Freyr Sverrisson, et~al.
\newblock De novo design of protein interactions with learned surface fingerprints.
\newblock \emph{Nature}, pp.\  1--9, 2023.

\bibitem[Gupta(2008)]{gupta2008metallo}
Varsha Gupta.
\newblock Metallo beta lactamases in pseudomonas aeruginosa and acinetobacter species.
\newblock \emph{Expert opinion on investigational drugs}, 17\penalty0 (2):\penalty0 131--143, 2008.

\bibitem[Hekkelman et~al.(2023)Hekkelman, de~Vries, Joosten, and Perrakis]{hekkelman2023alphafill}
Maarten~L Hekkelman, Ida de~Vries, Robbie~P Joosten, and Anastassis Perrakis.
\newblock Alphafill: enriching alphafold models with ligands and cofactors.
\newblock \emph{Nature Methods}, 20\penalty0 (2):\penalty0 205--213, 2023.

\bibitem[Hoffman et~al.(2022)Hoffman, Chenthamarakshan, Wadhawan, Chen, and Das]{hoffman2022optimizing}
Samuel~C Hoffman, Vijil Chenthamarakshan, Kahini Wadhawan, Pin-Yu Chen, and Payel Das.
\newblock Optimizing molecules using efficient queries from property evaluations.
\newblock \emph{Nature Machine Intelligence}, 4\penalty0 (1):\penalty0 21--31, 2022.

\bibitem[Hsu et~al.(2022)Hsu, Verkuil, Liu, Lin, Hie, Sercu, Lerer, and Rives]{hsu2022learning}
Chloe Hsu, Robert Verkuil, Jason Liu, Zeming Lin, Brian Hie, Tom Sercu, Adam Lerer, and Alexander Rives.
\newblock Learning inverse folding from millions of predicted structures.
\newblock In \emph{International Conference on Machine Learning}, pp.\  8946--8970. PMLR, 2022.

\bibitem[Ingraham et~al.(2019)Ingraham, Garg, Barzilay, and Jaakkola]{ingraham2019generative}
John Ingraham, Vikas Garg, Regina Barzilay, and Tommi Jaakkola.
\newblock Generative models for graph-based protein design.
\newblock \emph{Advances in neural information processing systems}, 32, 2019.

\bibitem[Jain et~al.(2022)Jain, Bengio, Hernandez-Garcia, Rector-Brooks, Dossou, Ekbote, Fu, Zhang, Kilgour, Zhang, et~al.]{jain2022biological}
Moksh Jain, Emmanuel Bengio, Alex Hernandez-Garcia, Jarrid Rector-Brooks, Bonaventure~FP Dossou, Chanakya~Ajit Ekbote, Jie Fu, Tianyu Zhang, Michael Kilgour, Dinghuai Zhang, et~al.
\newblock Biological sequence design with gflownets.
\newblock In \emph{International Conference on Machine Learning}, pp.\  9786--9801. PMLR, 2022.

\bibitem[Jing et~al.(2020)Jing, Eismann, Suriana, Townshend, and Dror]{jing2020learning}
Bowen Jing, Stephan Eismann, Patricia Suriana, Raphael~JL Townshend, and Ron Dror.
\newblock Learning from protein structure with geometric vector perceptrons.
\newblock \emph{arXiv preprint arXiv:2009.01411}, 2020.

\bibitem[Kumar \& Levine(2020)Kumar and Levine]{kumar2020model}
Aviral Kumar and Sergey Levine.
\newblock Model inversion networks for model-based optimization.
\newblock \emph{Advances in Neural Information Processing Systems}, 33:\penalty0 5126--5137, 2020.

\bibitem[Lin \& AlQuraishi(2023)Lin and AlQuraishi]{lin2023generating}
Yeqing Lin and Mohammed AlQuraishi.
\newblock Generating novel, designable, and diverse protein structures by equivariantly diffusing oriented residue clouds.
\newblock \emph{arXiv preprint arXiv:2301.12485}, 2023.

\bibitem[Lin et~al.(2022)Lin, Akin, Rao, Hie, Zhu, Lu, Smetanin, dos Santos~Costa, Fazel-Zarandi, Sercu, Candido, et~al.]{lin2022language}
Zeming Lin, Halil Akin, Roshan Rao, Brian Hie, Zhongkai Zhu, Wenting Lu, Nikita Smetanin, Allan dos Santos~Costa, Maryam Fazel-Zarandi, Tom Sercu, Sal Candido, et~al.
\newblock Language models of protein sequences at the scale of evolution enable accurate structure prediction.
\newblock \emph{bioRxiv}, 2022.

\bibitem[Madani et~al.(2020)Madani, McCann, Naik, Keskar, Anand, Eguchi, Huang, and Socher]{madani2020progen}
Ali Madani, Bryan McCann, Nikhil Naik, Nitish~Shirish Keskar, Namrata Anand, Raphael~R Eguchi, Po-Ssu Huang, and Richard Socher.
\newblock Progen: Language modeling for protein generation.
\newblock \emph{arXiv preprint arXiv:2004.03497}, 2020.

\bibitem[McNutt et~al.(2021)McNutt, Francoeur, Aggarwal, Masuda, Meli, Ragoza, Sunseri, and Koes]{mcnutt2021gnina}
Andrew~T McNutt, Paul Francoeur, Rishal Aggarwal, Tomohide Masuda, Rocco Meli, Matthew Ragoza, Jocelyn Sunseri, and David~Ryan Koes.
\newblock Gnina 1.0: molecular docking with deep learning.
\newblock \emph{Journal of cheminformatics}, 13\penalty0 (1):\penalty0 1--20, 2021.

\bibitem[McPartlon et~al.(2022)McPartlon, Lai, and Xu]{mcpartlon2022deep}
Matt McPartlon, Ben Lai, and Jinbo Xu.
\newblock A deep se (3)-equivariant model for learning inverse protein folding.
\newblock \emph{bioRxiv}, pp.\  2022--04, 2022.

\bibitem[Melnyk et~al.(2021)Melnyk, Das, Chenthamarakshan, and Lozano]{melnyk2021benchmarking}
Igor Melnyk, Payel Das, Vijil Chenthamarakshan, and Aurelie Lozano.
\newblock Benchmarking deep generative models for diverse antibody sequence design.
\newblock \emph{arXiv preprint arXiv:2111.06801}, 2021.

\bibitem[Moss et~al.(2020)Moss, Leslie, Beck, Gonzalez, and Rayson]{moss2020boss}
Henry Moss, David Leslie, Daniel Beck, Javier Gonzalez, and Paul Rayson.
\newblock Boss: Bayesian optimization over string spaces.
\newblock \emph{Advances in neural information processing systems}, 33:\penalty0 15476--15486, 2020.

\bibitem[Packer \& Liu(2015)Packer and Liu]{packer2015methods}
Michael~S Packer and David~R Liu.
\newblock Methods for the directed evolution of proteins.
\newblock \emph{Nature Reviews Genetics}, 16\penalty0 (7):\penalty0 379--394, 2015.

\bibitem[Palzkill(2013)]{palzkill2013metallo}
Timothy Palzkill.
\newblock Metallo-$\beta$-lactamase structure and function.
\newblock \emph{Annals of the New York Academy of Sciences}, 1277\penalty0 (1):\penalty0 91--104, 2013.

\bibitem[Park et~al.(2006)Park, Nam, Lee, Yoon, Mannervik, Benkovic, and Kim]{park2006design}
Hee-Sung Park, Sung-Hun Nam, Jin~Kak Lee, Chang~No Yoon, Bengt Mannervik, Stephen~J Benkovic, and Hak-Sung Kim.
\newblock Design and evolution of new catalytic activity with an existing protein scaffold.
\newblock \emph{Science}, 311\penalty0 (5760):\penalty0 535--538, 2006.

\bibitem[Ren et~al.(2022)Ren, Li, Ding, Zhou, Ma, and Peng]{ren2022proximal}
Zhizhou Ren, Jiahan Li, Fan Ding, Yuan Zhou, Jianzhu Ma, and Jian Peng.
\newblock Proximal exploration for model-guided protein sequence design.
\newblock \emph{bioRxiv}, 2022.

\bibitem[Richter et~al.(2011)Richter, Leaver-Fay, Khare, Bjelic, and Baker]{richter2011novo}
Florian Richter, Andrew Leaver-Fay, Sagar~D Khare, Sinisa Bjelic, and David Baker.
\newblock De novo enzyme design using rosetta3.
\newblock \emph{PloS one}, 6\penalty0 (5):\penalty0 e19230, 2011.

\bibitem[Sasportas et~al.(2009)Sasportas, Kasmieh, Wakimoto, Hingtgen, van~de Water, Mohapatra, Figueiredo, Martuza, Weissleder, and Shah]{sasportas2009assessment}
Laura~S Sasportas, Randa Kasmieh, Hiroaki Wakimoto, Shawn Hingtgen, Jeroen~AJM van~de Water, Gayatry Mohapatra, Jose~Luiz Figueiredo, Robert~L Martuza, Ralph Weissleder, and Khalid Shah.
\newblock Assessment of therapeutic efficacy and fate of engineered human mesenchymal stem cells for cancer therapy.
\newblock \emph{Proceedings of the national academy of sciences}, 106\penalty0 (12):\penalty0 4822--4827, 2009.

\bibitem[Shi et~al.(2022)Shi, Wang, Lu, Zhong, and Tang]{shi2022protein}
Chence Shi, Chuanrui Wang, Jiarui Lu, Bozitao Zhong, and Jian Tang.
\newblock Protein sequence and structure co-design with equivariant translation.
\newblock \emph{arXiv preprint arXiv:2210.08761}, 2022.

\bibitem[Song \& Li(2023)Song and Li]{song2023importance}
Zhenqiao Song and Lei Li.
\newblock Importance weighted expectation-maximization for protein sequence design.
\newblock \emph{arXiv preprint arXiv:2305.00386}, 2023.

\bibitem[Springer et~al.(1994)Springer, Sligar, Olson, and Phillips]{springer1994mechanisms}
Barry~A Springer, Stephen~G Sligar, John~S Olson, and George N~Jr Phillips.
\newblock Mechanisms of ligand recognition in myoglobin.
\newblock \emph{Chemical Reviews}, 94\penalty0 (3):\penalty0 699--714, 1994.

\bibitem[Terayama et~al.(2021)Terayama, Sumita, Tamura, and Tsuda]{terayama2021black}
Kei Terayama, Masato Sumita, Ryo Tamura, and Koji Tsuda.
\newblock Black-box optimization for automated discovery.
\newblock \emph{Accounts of Chemical Research}, 54\penalty0 (6):\penalty0 1334--1346, 2021.

\bibitem[Trippe et~al.(2022)Trippe, Yim, Tischer, Baker, Broderick, Barzilay, and Jaakkola]{trippe2022diffusion}
Brian~L Trippe, Jason Yim, Doug Tischer, David Baker, Tamara Broderick, Regina Barzilay, and Tommi~S Jaakkola.
\newblock Diffusion probabilistic modeling of protein backbones in 3d for the motif-scaffolding problem.
\newblock In \emph{The Eleventh International Conference on Learning Representations}, 2022.

\bibitem[Vaswani et~al.(2017)Vaswani, Shazeer, Parmar, Uszkoreit, Jones, Gomez, Kaiser, and Polosukhin]{vaswani2017attention}
Ashish Vaswani, Noam Shazeer, Niki Parmar, Jakob Uszkoreit, Llion Jones, Aidan~N Gomez, {\L}ukasz Kaiser, and Illia Polosukhin.
\newblock Attention is all you need.
\newblock \emph{Advances in neural information processing systems}, 30, 2017.

\bibitem[Wang et~al.(2022)Wang, Lisanza, Juergens, Tischer, Watson, Castro, Ragotte, Saragovi, Milles, Baek, et~al.]{wang2022scaffolding}
Jue Wang, Sidney Lisanza, David Juergens, Doug Tischer, Joseph~L Watson, Karla~M Castro, Robert Ragotte, Amijai Saragovi, Lukas~F Milles, Minkyung Baek, et~al.
\newblock Scaffolding protein functional sites using deep learning.
\newblock \emph{Science}, 377\penalty0 (6604):\penalty0 387--394, 2022.

\bibitem[Watson et~al.(2023)Watson, Juergens, Bennett, Trippe, Yim, Eisenach, Ahern, Borst, Ragotte, Milles, et~al.]{watson2023novo}
Joseph~L Watson, David Juergens, Nathaniel~R Bennett, Brian~L Trippe, Jason Yim, Helen~E Eisenach, Woody Ahern, Andrew~J Borst, Robert~J Ragotte, Lukas~F Milles, et~al.
\newblock De novo design of protein structure and function with rfdiffusion.
\newblock \emph{Nature}, pp.\  1--3, 2023.

\bibitem[Xiong et~al.(2020)Xiong, Hu, Huang, Zhang, Chen, and Liu]{xiong2020increasing}
Peng Xiong, Xiuhong Hu, Bin Huang, Jiahai Zhang, Quan Chen, and Haiyan Liu.
\newblock Increasing the efficiency and accuracy of the abacus protein sequence design method.
\newblock \emph{Bioinformatics}, 36\penalty0 (1):\penalty0 136--144, 2020.

\bibitem[Yim et~al.(2023)Yim, Trippe, De~Bortoli, Mathieu, Doucet, Barzilay, and Jaakkola]{yim2023se}
Jason Yim, Brian~L Trippe, Valentin De~Bortoli, Emile Mathieu, Arnaud Doucet, Regina Barzilay, and Tommi Jaakkola.
\newblock Se (3) diffusion model with application to protein backbone generation.
\newblock \emph{arXiv preprint arXiv:2302.02277}, 2023.

\end{thebibliography}
